\documentclass[10pt,twocolumn,letterpaper]{article}

\usepackage{cvpr}              % To produce the CAMERA-READY version
\usepackage{graphicx}
\usepackage{amsmath}
\usepackage{amssymb}
\usepackage{booktabs}
\usepackage{multirow}
\usepackage{tabularx}
\usepackage{pifont}
\usepackage[misc]{ifsym}
\usepackage{amsmath}
\usepackage{amssymb}
\usepackage{mathtools}
\usepackage{amsthm}

\usepackage[most]{tcolorbox} % 核心宏包
\newtcolorbox{mybox}[1]{
    enhanced,
    boxrule=1pt,            % 核心修改：将最外圈线条设为极细
    titlerule=1pt,          % 标题下方的横线也设为同等粗细
    colback=white,
    colframe=blue!30!black,
    fonttitle=\bfseries,
    coltitle=black,
    colbacktitle=blue!10!white, % 标题栏浅蓝色背景
    arc=4pt,
    title=#1,
    titlerule=0.5pt,
    titlerule style=blue!30!black,
    left=10pt,
    right=10pt,
    top=8pt,
    bottom=8pt,
}
\usepackage{enumitem}
\usepackage{booktabs}    % 核心：提供三线表命令
\usepackage{multirow}    % 用于垂直合并单元格
\usepackage{tabularx}    % 用于自动调整列宽
\usepackage{color}
\usepackage{bm}

\newcommand{\tf}[1]{{\color{red} [ToFill]}}
\newcommand{\tc}[1]{{\color{red} [ToCite]}}

\makeatletter
\def\blfootnote{\xdef\@thefnmark{}\@footnotetext}
\makeatother

\usepackage[pagebackref,breaklinks,colorlinks]{hyperref}

\usepackage[capitalize]{cleveref}
\crefname{section}{Sec.}{Secs.}
\Crefname{section}{Section}{Sections}
\Crefname{table}{Table}{Tables}
\crefname{table}{Tab.}{Tabs.}

\begin{document}

\title{Audit After Segmentation: Reference-Free Mask Quality Assessment for Language-Referred Audio-Visual Segmentation}

\author{Jinxing Zhou$^{1}$, Yanghao Zhou$^2$, Yaoting Wang$^3$, Zongyan Han$^1$, Jiaqi Ma$^1$, \\
Henghui Ding$^3$, Rao Muhammad Anwer$^1$, Hisham Cholakkal$^1$
\vspace{2mm} \\
$^{1}$MBZUAI, 
$^{2}$National University of Singapore,
$^{3}$Fudan University
}
\maketitle

\blfootnote{ $^\textrm{\Letter}$Correspondence to \textit{jinxing.zhou@mbzuai.ac.ae}.}

%%%%%%%%% ABSTRACT
% \vspace{3mm}

\begin{abstract}
Language-referred audio-visual segmentation (Ref-AVS) aims to segment target objects described by natural language by jointly reasoning over video, audio, and text.
Beyond generating segmentation masks, providing rich and interpretable diagnoses of mask quality remains largely underexplored.
In this work, we introduce Mask Quality Assessment in the Ref-AVS context (MQA-RefAVS), a new task that evaluates the quality of candidate segmentation masks without relying on ground-truth annotations as references at inference time.
Given audio-visual-language inputs and each provided segmentation mask, the task requires estimating its IoU with the unobserved ground truth, identifying the corresponding error type, and recommending an actionable quality-control decision.
To support this task, we construct MQ-RAVSBench, a benchmark featuring diverse and representative mask error modes that span both geometric and semantic issues.
We further propose MQ-Auditor, a multimodal large language model (MLLM)-based auditor that explicitly reasons over multimodal cues and mask information to produce quantitative and qualitative mask quality assessments.
Extensive experiments demonstrate that MQ-Auditor outperforms strong open-source and commercial MLLMs and can be integrated with existing Ref-AVS systems to detect segmentation failures and support downstream segmentation improvement. 
Data and codes will be released at \url{https://github.com/jasongief/MQA-RefAVS}.

\end{abstract}

\section{Introduction}
Object segmentation has long been a fundamental problem in computer vision, evolving from image segmentation~\cite{minaee2021image} to video object segmentation~\cite{zhou2022survey}. 
With the introduction of external textual and audio guidance, research has further extended to reference-guided video object segmentation~\cite{ding2025multimodal} and audio-guided visual segmentation~\cite{li2025waveforms}.
Building upon them, \textit{language-referred audio-visual segmentation} (Ref-AVS)~\cite{wang2024ref} has been recently proposed and emerged as a popular research topic in multimodal segmentation field.
Given a video with synchronized audio and a reference text, the Ref-AVS task seeks to produce segmentation masks that accurately ground the referred object in audio-visual scenes.

\begin{figure}[t]
  \centering
\includegraphics[width=\columnwidth]{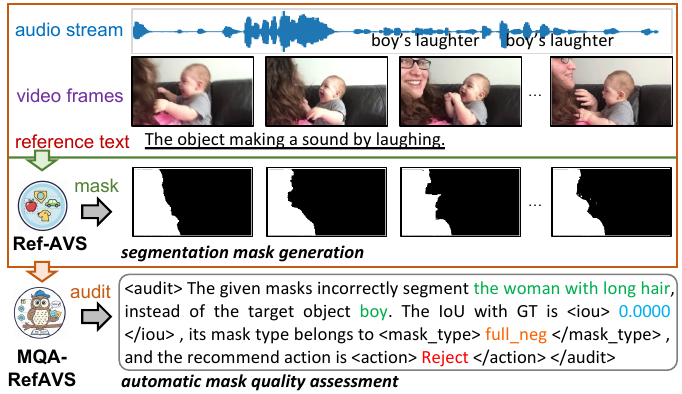}
\vspace{-4ex}
   \caption{Task illustration. Prior Ref-AVS methods aim to segment the target object. In contrast, our proposed MQA-RefAVS task focuses on automatic mask quality assessment, enabling to identify mask errors and provide suitable actions for further refinement.}
   \label{fig:intro}
% \vspace{-4ex}
\end{figure}

Current Ref-AVS research overwhelmingly focuses on generating segmentation masks and primarily uses these masks to evaluate model performance by computing the Intersection over Union (IoU) against ground-truth masks.
However, ground-truth masks are often unavailable in practical deployment (\textit{i.e., reference-free}).
For example, when constructing the seminal Ref-AVSBench dataset~\cite{wang2024ref} for the Ref-AVS task or evaluating a Ref-AVS segmentation model on new datasets, the ground-truth masks of the target data are unavailable.
In such situations, an auditor (human or model) must not only interpret the audio, video, and reference text to identify the target object, but also carefully judge whether a pre-annotated or generated mask truly corresponds to the intended object and is reliable for downstream use.
This process fundamentally differs from segmentation itself and thus constitutes a distinct problem: \textbf{Mask Quality Assessment (MQA)}.
In practice, segmentation systems inevitably produce masks with diverse failure modes.
Fig.~\ref{fig:intro} shows an example in which a prior state-of-the-art Ref-AVS model, TGS-Agent~\cite{zhou2025think}, fails to reason over multimodal content and incorrectly segments the \textit{woman} instead of the target \textit{boy}.
Previously, we could only rely on ground-truth masks to obtain a single IoU score.
% By contrast to scalar evaluation metrics, MQA is able to provide a richer and more interpretable mask quality assessment in natural language.
By contrast to scalar evaluation metrics that only summarize performance, MQA provides richer and more interpretable assessments (see Fig.~\ref{fig:intro}) that support fine-grained quality control.
After identifying quality issues in the generated masks, MQA can support mask re-generation or targeted revision, thereby facilitating more reliable segmentation pipelines.
However, existing datasets and models provide little support for explicitly modeling or automating this step.

To fill this gap, we introduce Mask Quality Assessment under the Ref-AVS context (\textbf{MQA-RefAVS}), a new task that aims to \textit{automatically infer the quality of candidate segmentation masks without access to ground-truth annotations}.
As shown in Fig.~\ref{fig:intro}, given multimodal inputs and each frame's predicted mask, this task requires estimating the mask IoU, identifying its error type, and recommending an appropriate action for quality control.
% To support this task,
To systematically study MQA under diverse and controllable failure patterns, we construct \textbf{MQ-RAVSBench}, the first benchmark specifically designed for mask quality assessment in Ref-AVS.
MQ-RAVSBench is built upon the Ref-AVSBench dataset~\cite{wang2024ref} and contains 1,840 videos with 2,046 reference texts, from which we generate 26,061 mask instances.
For each $<$video, reference$>$ pair, we generate six representative mask types, including \textit{perfect}, \textit{cutout}, \textit{dilate}, \textit{erode}, \textit{merge}, and \textit{full\_neg}, covering diverse failure modes.
Each mask is annotated with its IoU and a recommended action selected from \textit{accept}, \textit{minor revision}, \textit{major revision}, and \textit{reject}.
These masks collectively mimic realistic quality variations, ranging from entirely accurate predictions to minor \textit{geometric} imperfections and severe \textit{semantic} errors.
All masks are automatically constructed using open-source multimodal models and tools, with detailed procedures provided in Sec.~\ref{sec:dataset}.

Furthermore, we propose \textbf{MQ-Auditor}, a multimodal large language model (MLLM)-based auditor that explicitly reasons over audio, visual, language, and mask information.
MQ-Auditor is trained via supervised instruction tuning.
Unlike prior approaches, MQ-Auditor assesses mask quality and estimates IoU without requiring ground-truth masks during inference, making it directly applicable at deployment time.
Extensive experiments on MQ-RAVSBench demonstrate that mask quality assessment, particularly in the multimodal Ref-AVS setting, remains non-trivial for general-purpose open-source and commercial MLLMs, including Gemini-3-Flash~\cite{google2025gemini3}. In contrast, MQ-Auditor consistently provides more accurate and reliable quality assessments. Moreover, MQ-Auditor can effectively complement existing Ref-AVS models by identifying segmentation failures and improving segmentation performance.

In summary, our contributions are threefold:
\textbf{1)} We identify mask quality assessment as a new and essential problem in language-referred audio-visual segmentation.
\textbf{2)} We establish MQ-RAVSBench, the first dataset for systematic mask quality assessment.
\textbf{3)} We propose MQ-Auditor, a multimodal auditor that infers mask quality without ground-truth and supports downstream segmentation improvement.

\section{Task: MQA-RefAVS}\label{sec:task_definition}
We explore mask quality assessment (MQA) within a representative multimodal segmentation task: the Ref-AVS setting.
Specifically, given a video $\mathcal{V}$ with synchronized audio $\mathcal{A}$, a referring expression $\mathcal{R}$ describing a target object, a key video frame $\mathcal{V}_t$ at the $t$-th segment ($t \in [1, T]$), and its corresponding candidate binary mask $\mathcal{M}_t$, the MQA-RefAVS task requires an \textit{auditor} model $\Phi$ to assess mask quality by predicting:
\textbf{1) the Intersection over Union (IoU) $\bm{s}$} between the candidate mask and the unobserved ground-truth mask, which serves as a quantitative quality measure with $s \in [0, 1]$;
\textbf{2) the mask type $\bm{m}$}, selected from six predefined categories: \textit{perfect, full\_neg, cutout, dilate, erode, merge};
and \textbf{3) the recommended action $\bm{a}$} for quality control, chosen from \textit{accept, minor revision, major revision, reject}.
Details of the mask type and action sets are introduced in Sec.~\ref{sec:mask_iou_action}.
For simplicity, we omit the subscript $t$ for $s$, $m$, and $a$.
In summary, the MQA-RefAVS task can be formulated as:
\begin{equation}
   s, m, a = \Phi(\mathcal{V}, \mathcal{A}, \mathcal{R}, \mathcal{V}_t, \mathcal{M}_t).
   \label{eq:task}
\end{equation}
This task requires jointly perceiving audio, video, and referring language to identify the target object, determine whether it is accurately segmented by the candidate mask in the given frame, and ultimately produce both quantitative and qualitative assessments.

\begin{figure*}[t]
  \centering
\includegraphics[width=1\textwidth]{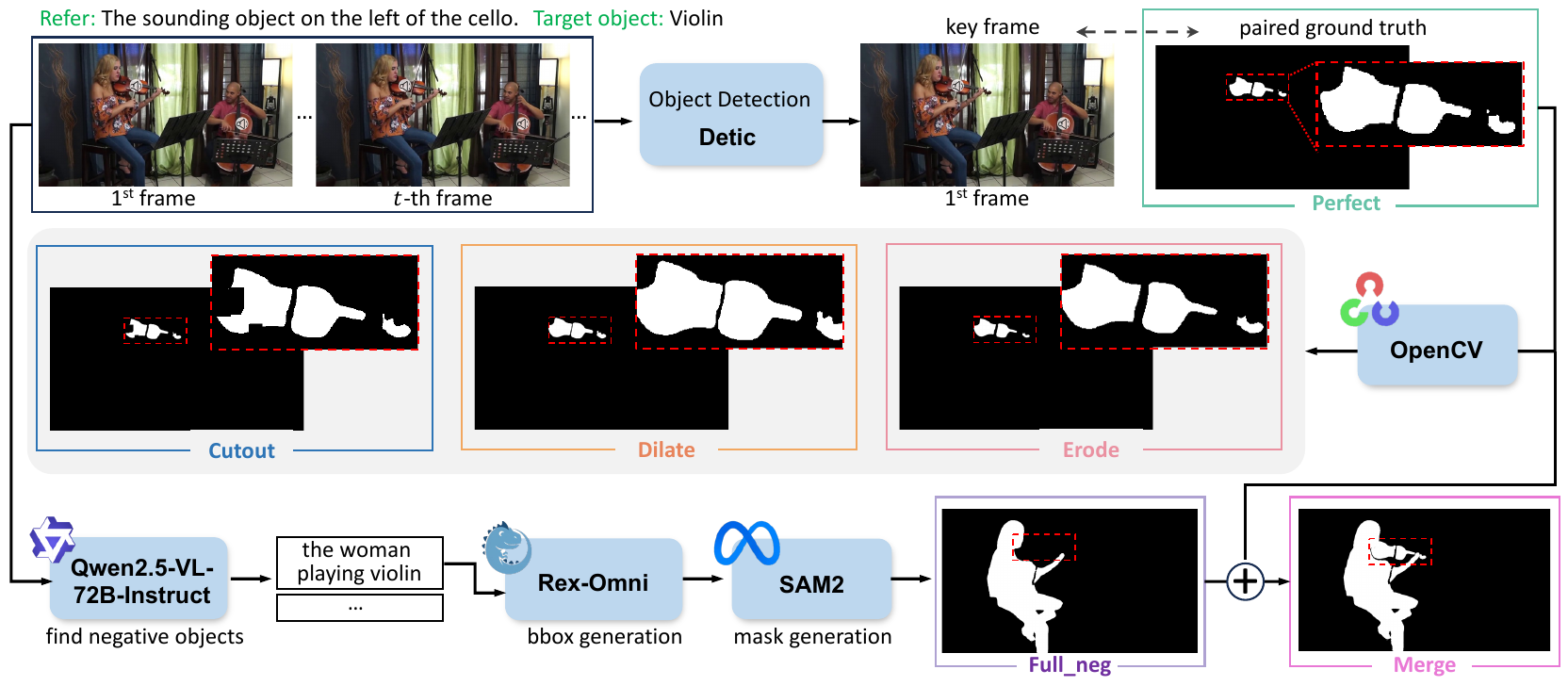}
\vspace{-4.5ex}
   \caption{Mask construction pipeline of MQ-RAVSBench.
    For training and image-based evaluation, we employ an object detection model, Detic~\cite{zhou2022detecting}, to identify a key frame containing the richest set of objects.
    Based on the ground-truth \textbf{Perfect} masks from Ref-AVSBench~\cite{wang2024ref}, we use OpenCV library to generate masks with \textit{geometric} quality issues, including \textbf{Cutout}, \textbf{Dilate}, and \textbf{Erode}.
    Besides, we construct a pipeline using powerful MLLMs/VLMs to generate \textbf{Full\_neg} masks, which correspond to entirely incorrect objects and exhibit severe \textit{semantic} quality issues.
    By combining \textit{Full\_neg} and \textit{Perfect} masks, we obtain the \textbf{Merge} masks.  % The final numbers of \textit{Perfect}, \textit{Cutout/Dilate/Erode}, and \textit{Merge/Full\_neg} masks are 1, 2, and up to 3 generated by controlling resulting IoU values.
   }
   \label{fig:mask_pipeline}
% \vspace{-2.5ex}
\end{figure*}

\section{Dataset: MQ-RAVSBench}\label{sec:dataset}
To facilitate the proposed task, we propose the MQ-RAVSBench. We introduce its construction details below.

\subsection{Data Source and Split}
We source videos from the existing Ref-AVSBench dataset~\cite{wang2024ref}, originally proposed for the Ref-AVS task.
Videos in Ref-AVSBench are 10 seconds long and are each associated with multiple referring texts, potentially targeting different objects.
For each $<$video, reference$>$ pair, the object category of the referring expression and binary segmentation masks for 10 uniformly sampled video frames are provided.
These metadata are crucial for our dataset construction.
During video sampling, we carefully balance video uniqueness, referring text templates, and object categories.
As a result, MQ-RAVSBench consists of 1,840 videos, which are split into 1,306 videos for training and 534 for testing.
To facilitate future evaluation of open-vocabulary and zero-shot generalization, the test set is further divided into two subsets: 269 videos in the \textit{Seen} set and 265 videos in the \textit{Unseen} set, depending on whether the object categories appear in the training set.
Each training video is paired with one referring text, yielding 1,306 $<$video, reference$>$ instances.
For the test set, each video may be associated with one or two referring texts, resulting in 437 and 303 instances for the Seen and Unseen subsets, respectively.
Notably, each training and testing instance is augmented with multiple masks (7$\sim$13) of varying quality levels (see Sec.~\ref{sec:mask_iou_action}), substantially expanding the overall dataset scale (see Table~\ref{tab:MQ-RAVSBench}).

\subsection{Mask Taxonomy and Quality Annotation}\label{sec:mask_iou_action}
After identifying the video and referring text pairs, we further determine the corresponding segmentation masks with diverse quality levels, compute the IoU values, and define recommended actions.
Specifically, we consider six mask types that mimic common cases encountered in human annotation and real-world model predictions.
An overview of the construction pipeline is shown in Fig.~\ref{fig:mask_pipeline}.

\noindent\textbf{Entirely Accurate (Perfect).}
The candidate mask $\mathcal{M}_t$ accurately segments the target referred object, precisely delineating its shape with all true positive pixels, and perfectly matches the ground-truth mask $\mathcal{G}_t$.
This type of mask is directly sourced from Ref-AVSBench~\cite{wang2024ref}, where each instance contains one perfect mask.
Accordingly, the IoU value $s$ of such a mask is 1, and the recommended action $a$ is \textit{accept}:
\begin{equation}
    \mathcal{M}_t = \mathcal{G}_t \boldsymbol{\Rightarrow} m=\{\text{perfect}\}, s = 1, a = \{\text{accept}\}.
\end{equation}

\noindent\textbf{Entirely Incorrect (Full\_neg).}
In this case, the candidate mask $\mathcal{M}_t$ is completely incorrect with respect to the ground truth, segmenting an entirely irrelevant object or background region.
Such errors are common when a human annotator or a segmentation model misinterprets the audio-visual-language cues and selects an incorrect target.
All pixels in $\mathcal{M}_t$ are false positives, resulting in an IoU value of 0, and the recommended action is \textit{reject}:
\begin{equation}
    \mathcal{M}_t \cap \mathcal{G}_t = 0 \boldsymbol{\Rightarrow} m=\{\text{full\_neg}\}, s = 0, a = \{\text{reject}\}.
\end{equation}
We generate full negative masks using an automated pipeline.
Given the known target object category for each $<$video, reference$>$ instance and a corresponding video frame $\mathcal{V}_t$, we employ a powerful vision--language model, Qwen2.5-VL-72B-Instruct-AWQ~\cite{bai2025qwen25vl}, to generate up to five negative object candidates.
Instead of simple object categories, the model outputs descriptive noun phrases, which more precisely distinguish visually similar objects.
{The detailed prompt is provided in Sec.~\ref{sec:prompt_for_full_neg}.}
Using these negative object phrases, we leverage Rex-Omni~\cite{jiang2025detect}, a state-of-the-art MLLM supporting referring expression grounding, to localize corresponding bounding boxes.
These bounding boxes are then used as prompts for SAM2~\cite{ravi2024sam2} to generate segmentation masks.
To further select challenging negatives, we compute the bounding-box IoU between the remaining negative masks and the ground-truth mask, and retain the top three masks with the highest overlap.
A higher bounding-box IoU indicates that the negative object is spatially closer to or interacts with the target object, making the resulting masks more challenging for multimodal understanding and quality assessment.
As shown in Fig.~\ref{fig:mask_pipeline}, the target object in the example is the \textit{violin}, while the generated negative object is the \textit{woman playing violin}.
Notably, in simpler scenarios, the number of valid full-negative masks may be fewer than three, ranging from zero to three.

\noindent\textbf{Internal Cutout (Cutout).}
In this case, the candidate mask $\mathcal{M}_t$ successfully localizes the target object but misses a subset of positive pixels $\mathcal{C}_t$ in its interior regions (Fig.~\ref{fig:mask_pipeline}).
Such errors may arise from the limited capacity of segmentation models (see Figs.~8 and 9 in prior work~\cite{zhou2022avs}).
In addition, segmentation models typically apply a fixed threshold to obtain final masks, and variations in thresholding can also lead to the removal of pixels in different regions.
Considering these factors, we generate two cutout masks by constraining the resulting IoU to the ranges $[0.85, 0.9]$ and $[0.75, 0.8]$, respectively.
These ranges are empirically determined in preliminary experiments to produce masks that are both realistic and challenging for quality assessment.
The mask with the higher IoU is regarded as a \textit{hard} sample, where cutout errors are more subtle and difficult to evaluate, while the lower IoU range corresponds to \textit{medium} difficulty.
Accordingly, the recommended actions for the hard and medium samples are defined as \textit{minor revision} and \textit{major revision}, respectively.
Based on the ground-truth \textit{perfect} masks provided by Ref-AVSBench, we utilize the OpenCV\footnote{https://github.com/opencv/opencv} library to generate \textit{cutout} masks using rectangular or elliptical structuring elements of varying sizes.
This process can be summarized as:
\begin{multline}
    \mathcal{M}_t = \mathcal{G}_t \setminus \mathcal{C}_t, \mathcal{C}_t \subset \mathcal{G}_t \boldsymbol{\Rightarrow}  m=\{\text{cutout}\},\\
    s \in (0,1), a = \{\text{minor revision; major revision}\}.
\end{multline}

\noindent\textbf{Local Dilation (Dilate).}
Although the candidate mask $\mathcal{M}_t$ successfully covers all pixels of the target object, its boundaries incorrectly expand into neighboring regions $\mathcal{D}_t$, resulting in outward over-segmentation that includes background pixels.
Similar to the \textit{cutout} mask type, we generate two dilation masks with \textit{hard} and \textit{medium} difficulty levels, respectively.
The corresponding recommended actions are \textit{minor revision} and \textit{major revision}, formulated as:
\begin{multline}
    \mathcal{M}_t = \mathcal{G}_t \cup \mathcal{D}_t, \mathcal{D}_t \subset \text{neighbor}(\mathcal{G}_t)  \boldsymbol{\Rightarrow} m=\{\text{dilate}\}, \\
    s \in (0,1), a = \{\text{minor revision; major revision}\}.
\end{multline}

\noindent\textbf{Local Erosion (Erode).}
In this case, all pixels in the candidate mask $\mathcal{M}_t$ belong to the target object, but the mask fails to cover pixels near the object boundary $\mathcal{E}_t$, resulting in inward under-segmentation.
Following the same strategy as for the \textit{cutout} and \textit{dilate} types, we construct two erosion masks with \textit{hard} and \textit{medium} difficulty levels, corresponding to the \textit{minor revision} and \textit{major revision} actions:
\begin{multline}
    \mathcal{M}_t = \mathcal{G}_t \setminus \mathcal{E}_t, \mathcal{E}_t \subset \text{boundary}(\mathcal{G}_t)  \boldsymbol{\Rightarrow} m=\{\text{erode}\}, \\
    s \in (0,1), a = \{\text{minor revision; major revision}\}.
\end{multline}

\noindent\textbf{Merge with Non-target Objects (Merge).}
In this case, the candidate mask $\mathcal{M}_t$ correctly segments the target object but also incorrectly over-segments additional distracting objects $\mathcal{H}_t$.
In practice, we construct such masks by merging the \textit{perfect} mask with the \textit{full\_neg} mask introduced earlier.
As a result, each instance can yield up to three \textit{merge} masks.
We then compute the IoU of the resulting masks and define the recommended actions based on the IoU values.
Due to the incorporation of different negative objects, the IoU $s$ of a merged mask may lie in the range $(0, 1)$.
In our setting, we further divide this range into three cases:
\textbf{1)} if $s \in [0.9, 1)$, the correctly segmented target object occupies a large proportion of the merged mask and the impact of the negative object is limited; we define the recommended action as \textit{minor revision} (\textit{hard} sample);
\textbf{2)} if $s \in [0.75, 0.9)$, a larger portion of negative pixels is included and the recommended action is \textit{major revision} (\textit{medium-hard} sample);
\textbf{3)} if $s \in (0, 0.75)$, the recommended action is \textit{reject} (\textit{easy} sample).
We adopt relatively high thresholds (\textit{e.g.}, $0.75$) because even when the IoU of a merged mask is high, it may still incorrectly include a distinct object, indicating a severe semantic misunderstanding that should be explicitly penalized.
This process can be summarized as:
\begin{multline}
    \mathcal{M}_t = \mathcal{G}_t \cup \mathcal{H}_t, \mathcal{G}_t \cap \mathcal{H}_t = \varnothing  \boldsymbol{\Rightarrow} m=\{\text{merge}\}, \\
    s \in (0,1), a = \{\text{minor revision; major revision; reject}\}.
\end{multline}

\vspace{-2ex}
\underline{{In summary}}, we construct masks to reflect a wide range of common scenarios, including accurate segmentation (\textit{perfect}), under-segmentation (\textit{cutout, erode, full\_neg}), and over-segmentation (\textit{dilate, merge}).
These mask types jointly cover both geometric and semantic quality issues at varying levels, with diverse IoU distributions and corresponding quality-control actions.

\begin{table}[t]
\centering
\tiny
\caption{Statistics of MQ-RAVSBench. We show the distribution of videos, referring expressions, and mask samples across different training and testing splits. The \textit{image-based} and \textit{video-based} evaluations represent the testing is conducted on single key frame or all frames for each video.}
\vspace{-2.5ex}
\resizebox{\columnwidth}{!}{
% \begin{tabular}{l S[table-format=4.0] S[table-format=4.0] S[table-format=5.0]}
\begin{tabular}{lrrr}
\toprule
 & \textbf{\#videos} & \textbf{\#refer.} & \textbf{\#masks} \\ \midrule
Train & 1,306 & 1,306 & 16,761 \\ \midrule
\textit{\textbf{image-based evaluation}} & & & \\ 
Test (seen) & 269 & 437 & 5,609 \\
Test (unseen) & 265 & 303 & 3,691 \\
\multicolumn{1}{r}{total} & 534 & 740 & 9,300 \\ \midrule
\textit{\textbf{video-based evaluation}} & & & \\ 
Test (seen) & 60 & 60 & 7,674 \\
Test (unseen) & 40 & 40 & 4,966 \\
\multicolumn{1}{r}{total} & 100 & 100 & 12,640 \\ \bottomrule
\end{tabular}
}
\label{tab:MQ-RAVSBench}
\vspace{-6ex}
\end{table}

\subsection{Training and Evaluation Protocols}\label{sec:evaluation_details}

\noindent\textbf{Training Protocol.}
The training set contains 1,306 $<$video, reference$>$ instances.
In our setting, for each training instance, we select a single representative video frame to construct the six mask types following Sec.~\ref{sec:mask_iou_action}.
We employ a strong object detection model, Detic~\cite{zhou2022detecting}, to detect potential objects across 10 video frames sampled at 1 FPS, and select the frame containing the largest number of detected objects.
The rich visual content in such a key frame supports more diverse negative object selection when constructing \textit{full\_neg} and \textit{merge} masks.
As a result, the training set contains 16,761 mask samples.
This single-frame-based design improves training efficiency while allowing the model to learn from diverse distributions of object categories, mask types, IoU values, and action labels.
% This single-frame-based design improves training efficiency while preserving similar distributions of object categories, mask types, IoU values, and action labels compared to using full video sequences.

\noindent\textbf{Evaluation Protocol.}
As shown in Table~\ref{tab:MQ-RAVSBench}, we consider two evaluation protocols at test time:
\textit{\textbf{1) Image-based evaluation.}}
Similar to training phase, for each video, we evaluate the model's MQA performance on a single key frame.
The resulting numbers of masks for the Seen and Unseen test sets are 5,609 and 3,691, respectively.
This protocol evaluates the model's fundamental mask quality assessment ability, including multimodal semantic understanding, IoU estimation accuracy, and the correctness of recommended actions.
\textit{\textbf{2) Video-based evaluation.}}
In this protocol, evaluation is conducted on all 10 frames of each video.
For ease of analysis and less inference time cost, we select 60 and 40 videos from the Seen and Unseen test sets, respectively, and constrain all frames within a video to share the same mask type.
This setting yields 7,674 and 4,966 mask samples for testing.
Compared with the image-based protocol, the video-based evaluation further enables analysis of the model's temporal consistency in mask quality prediction.
Detailed statistics for each mask type are provided in Table~\ref{tab:data_stat_of_each_mask_type} in Sec.~\ref{sec:appendix_more_data_stat}.

\noindent\textbf{Evaluation Metrics.}
After defining training and testing protocols, we establish the evaluation metrics.
As introduced in Sec.~\ref{sec:task_definition}, the task requires predicting IoU $s$, mask type $m$, and recommended action $a$.
Accordingly, we design metrics to assess model performance on these aspects:\\
\textbullet{} \textbf{{RMSE}}.
This metric computes the Root Mean Square Error (RMSE) between the predicted IoU $s$ and the ground-truth IoU over all evaluation samples (see Eq.~\ref{eq:RMSE}).\\
% This metric computes the Root Mean Square Error (RMSE) between the predicted IoU $s$ and the ground-truth IoU $s_g$ over all $N$ evaluation samples:
% \begin{equation}
%     \text{RMSE} = \sqrt{ \frac{1}{N} \sum_{i=1}^{N} (s - s_g)^2 }.
% \end{equation}
\textbullet{} $\bm{F_2}$\textbf{-score}.
Both mask type and action predictions are evaluated using the ${F_\beta}$ score with $\beta = 2$ (see Eq.~\ref{eq:F2_score}).
This choice emphasizes \textit{recall}, which is desirable for MQA systems considering two factors:
{1)} missing a problematic mask is typically more costly than incorrectly flagging a correct one;
{2)} our empirical results show that MQA models tend to achieve high precision (often close to 100\%) but comparatively lower recall.
% The ${F_2}$-score is computed as:
% \begin{equation}
% {F_{\beta}} = (1 + \beta^2) \cdot \frac{\mathrm{Precision} \cdot \mathrm{Recall}}
% {\beta^2 \cdot \mathrm{Precision} + \mathrm{Recall}}.
% \end{equation}
More detailed calculations for each evaluation protocol are provided in Sec.~\ref{sec:supp_metric_details}.

\begin{figure}[t]
  \centering
\includegraphics[width=\columnwidth]{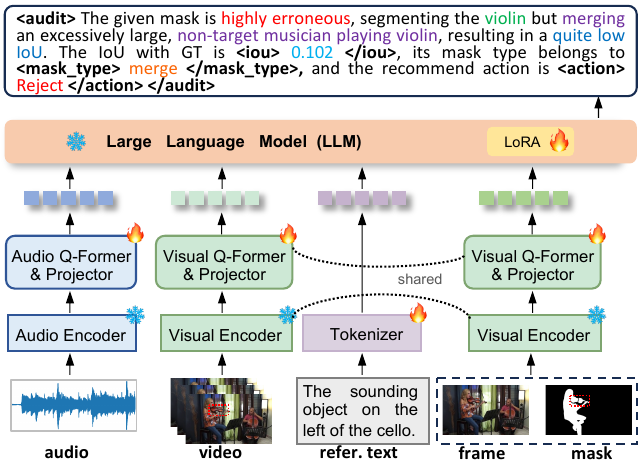}
\vspace{-4ex}
   \caption{Illustration of our MQ-Auditor model.}
   \label{fig:auditor_model}
\vspace{-2ex}
\end{figure}

\begin{table*}[t]
\centering
\caption{Image-based evaluation on MQ-RAVSBench.
We compare MQ-Auditor with strong open-source and commercial models, including Video-LLaMA3-7B~\cite{zhang2025videollama3}, Qwen2.5-Omni-7B~\cite{xu2025qwen25omni}, Ming-Flash-Omni~\cite{ai2025mingomni}, and Gemini-3-Flash-Preview~\cite{google2025gemini3}.
$F_2$-M (\%) and $F_2$-A (\%) denote the $F_2$-scores for mask type and action predictions, respectively.
`H' and `M' denote hard and medium-hard samples, respectively. `All' means that all mask samples of that type are evaluated.
{`Avg.'} denotes the average performance across all mask types.
The best and second-best results are highlighted in \textbf{bold} and \underline{underline}, respectively.}

\vspace{-1.5ex}

\small
\resizebox{1.0\textwidth}{!}{

\begin{tabular}{lll c cc cc cc c c c}
\toprule
\multirow{2.5}{*}{\textbf{Settings}} & \multirow{2.5}{*}{\textbf{Metrics}} & \multirow{2.5}{*}{\textbf{Methods}}  & \textbf{Perfect} & \multicolumn{2}{c}{\textbf{Cutout}} & \multicolumn{2}{c}{\textbf{Dilate}} & \multicolumn{2}{c}{\textbf{Erode}} & \textbf{Merge} & \textbf{Full\_neg} & \multirow{2.5}{*}{\textbf{Avg.}} \\ 
\cmidrule(lr){4-4} \cmidrule(lr){5-6} \cmidrule(lr){7-8} \cmidrule(lr){9-10} \cmidrule(lr){11-11} \cmidrule(lr){12-12}
 & & & All & H & M & H & M & H & M & All & All & \\ 
\midrule

\multirow{15}{*}{{Seen}} 
 & \multirow{5}{*}{RMSE $\downarrow$} 
 & Video-LLaMA3-7B & \textbf{0.249} & 0.268 & 0.291 & \underline{0.283} & 0.313 & \underline{0.272} & \underline{0.301} & 0.573 & 0.957 & 0.390  \\
 & & Qwen2.5-Omni-7B & 0.315 & 0.342 & 0.359 & 0.351 & 0.365 & 0.354 & 0.352 & 0.541 & 0.904 & 0.432 \\
 & &  Ming-Flash-Omni & 0.997 & 0.875 & 0.810 & 0.859 & 0.761 & 0.860 & 0.762 & 0.613 & 0.543 & 0.787 \\
 & & Gemini-3-Flash-Preview & 0.399 & \underline{0.363} & \underline{0.338} & 0.327 & \underline{0.297} & 0.347 & 0.308 & \textbf{0.304} & \textbf{0.251} & \underline{0.326} \\
 & & \textbf{MQ-Auditor (ours)} & \underline{0.359} & \textbf{0.199} & \textbf{0.160} & \textbf{0.298} & \textbf{0.232} & \textbf{0.329} & \textbf{0.305} & \underline{0.334} & \underline{0.507} & \textbf{0.303}   \\ 
\cmidrule{2-13}

 & \multirow{5}{*}{$F_2$-M $\uparrow$} 
 & Video-LLaMA3-7B & \underline{86.76} & 0.57 & 0.00 & 0.00 & 0.00 & 0.00 & 0.00 & 0.20 & 0.42 & 9.77 \\
 & & Qwen2.5-Omni-7B & \textbf{88.32} & 10.91 & 14.71 & 0.00 & 0.00 & 0.00 & 0.00 & 0.72 & 12.35 & 14.11 \\
 & &  Ming-Flash-Omni & 2.29 & 6.24 & 5.68 & 0.00 & 0.00 & 0.29 & 0.29 & 24.13 & 18.21 & 6.35 \\
 & & Gemini-3-Flash-Preview & 20.57 & \underline{77.78} & \underline{79.00} & \underline{0.57} & \underline{3.13} & \underline{3.41} & \underline{3.99} & \underline{73.98} & \textbf{80.40} & \underline{38.09} \\
 & & \textbf{MQ-Auditor (ours)} & 60.58 & \textbf{83.41} & \textbf{91.78} & \textbf{59.92} & \textbf{78.39} & \textbf{36.09} & \textbf{47.15} & \textbf{74.55} & \underline{69.43} & \textbf{66.81} \\ 
\cmidrule{2-13}

 & \multirow{5}{*}{$F_2$-A $\uparrow$} 
 & Video-LLaMA3-7B & \textbf{89.48} & 0.86 & 0.00 & 0.57 & 0.00 & 0.29 & 0.00 & 0.00 & 0.00 & 10.13   \\
 & & Qwen2.5-Omni-7B & \underline{88.12} & 11.46 & 0.00 & 7.05 & 0.00 & 5.66 & 0.00 & 4.96 & 12.35 & 14.40  \\
 & &  Ming-Flash-Omni & 1.43 & 0.86 & 4.84 & 1.43 & 3.71 & 0.57 & 3.99 & \underline{64.21} & 68.91 & 16.66 \\
 & & Gemini-3-Flash-Preview & 20.57 & \underline{20.04} & \underline{62.12} & \underline{23.97} & \underline{38.05} & \textbf{24.74} & \textbf{47.11} & \textbf{68.32} & \textbf{88.39} & \underline{43.70} \\
 & & \textbf{MQ-Auditor (ours)} & 60.58 & \textbf{29.35} & \textbf{85.78} & \textbf{24.74} & \textbf{71.57} & \underline{20.04} & \underline{44.08} & 43.00 & \underline{74.52} & \textbf{50.41} \\ 
\midrule

\multirow{15}{*}{{Unseen}} 
 & \multirow{5}{*}{RMSE $\downarrow$} 
 & Video-LLaMA3-7B & \textbf{0.192} & 0.259 & 0.306 & 0.264 & 0.298 & \underline{0.254} & 0.289 & 0.524 & 0.951 & 0.371 \\
 & & Qwen2.5-Omni-7B & 0.418 & 0.501 & 0.455 & 0.431 & 0.412 & 0.394 & 0.370 & 0.501 & 0.916 & 0.489 \\
 & &  Ming-Flash-Omni & 0.981 & 0.886 & 0.843 & 0.850 & 0.757 & 0.860 & 0.761 & 0.653 & 0.518 & 0.790 \\
 & & Gemini-3-Flash-Preview & \underline{0.409} & \underline{0.357} & \underline{0.334} & \underline{0.320} & \underline{0.298} & 0.345 & \underline{0.314} & \textbf{0.301} & \textbf{0.250} & \underline{0.325} \\
 & & \textbf{MQ-Auditor (ours)} & 0.274 & \textbf{0.161} & \textbf{0.135} & \textbf{0.203} & \textbf{0.154} & \textbf{0.257} & \textbf{0.269} & \underline{0.310} & \underline{0.508} & \textbf{0.252} \\ 
\cmidrule{2-13}

 & \multirow{5}{*}{$F_2$-M $\uparrow$} 
 & Video-LLaMA3-7B & \textbf{88.28} & 0.82 & 0.00 & 0.00 & 0.00 & 0.00 & 0.00 & 0.00 & 0.16 & 9.92  \\
 & & Qwen2.5-Omni-7B & \underline{74.74} & 24.71 & 26.19 & 0.00 & 0.00 & 0.00 & 0.00 & 1.13 & 3.71 & 14.50  \\
 & &  Ming-Flash-Omni & 5.52 & 15.85 & 14.60 & 0.00 & 0.00 & 0.43 & 1.71 & 26.63 & 17.64 & 9.15 \\
 & & Gemini-3-Flash-Preview & 18.39 & \underline{78.07} & \underline{79.40} & \underline{0.86} & \underline{3.16} & \underline{3.14} & \underline{3.97} & \underline{73.84} & \textbf{80.42} & \underline{37.92} \\
 & & \textbf{MQ-Auditor (ours)} & 64.34 & \textbf{93.29} & \textbf{97.07} & \textbf{76.22} & \textbf{89.43} & \textbf{27.30} & \textbf{41.26} & \textbf{77.73} & \underline{70.92} & \textbf{70.84} \\ 
\cmidrule{2-13}

 & \multirow{5}{*}{$F_2$-A $\uparrow$} 
 & Video-LLaMA3-7B & \textbf{94.09} & 0.00 & 0.00 & 0.00 & 0.00 & 0.00 & 0.00 & 0.00 & 0.16 & 10.47 \\
 & & Qwen2.5-Omni-7B & \underline{74.44} & \underline{25.08} & 0.00 & \underline{14.81} & 0.00 & \underline{13.64} & 0.00 & 6.94 & 3.71 & 15.40 \\
 & &  Ming-Flash-Omni & 5.94 & 1.36 & 11.34 & 2.99 & \underline{9.69} & 2.99 & \underline{8.86} & 57.88 & 66.20 & 18.58 \\
 & & Gemini-3-Flash-Preview & 21.98 & 19.78 & \underline{64.31} & 23.29 & 37.47 & \underline{24.85} & \textbf{47.15} & \textbf{69.08} & \textbf{88.49} & \underline{44.04} \\
 & & \textbf{MQ-Auditor (ours)} & 64.34 & \textbf{47.76} & \textbf{76.82} & \textbf{37.05} & \textbf{81.20} & \textbf{28.77} & 30.60 & \underline{44.57} & \underline{74.41} & \textbf{53.95} \\ 
\bottomrule

\end{tabular}
}
\vspace{-2ex}
\label{tab:image_based_evaluation_main}
\end{table*}

\section{Method: MQ-Auditor}\label{sec:model}
\noindent\textbf{Network.}
To address the MQA-RefAVS task, we propose a baseline approach named MQ-Auditor, which performs mask quality assessment by leveraging a multimodal large language model (MLLM) to analyze multimodal signals and produce judgments in natural language.
As shown in Fig.~\ref{fig:auditor_model}, the audio $\mathcal{A}$ and video $\mathcal{V}$ are processed by modality-specific encoders, BEATs~\cite{chen2022beats} and CLIP-ViT-L/14~\cite{radford2021CLIP}, respectively.
The extracted features are then passed through Q-Former modules~\cite{li2023blip} with 32 learnable query tokens, followed by linear projectors, to obtain audio and visual latent embeddings.
The referring expression $\mathcal{R}$ is tokenized and converted into text token embeddings.
Importantly, given a raw video frame $\mathcal{V}_t$ and a candidate mask $\mathcal{M}_t$, we first generate a masked frame $\mathcal{V}'_t$ by element-wise multiplication between $\mathcal{V}_t$ and $\mathcal{M}_t$.
Compared with the original frame $\mathcal{V}_t$, the masked frame $\mathcal{V}'_t$ explicitly highlights the regions selected by the mask.
The binary mask $\mathcal{M}_t$ is further converted into a pseudo-RGB image by channel duplication and concatenated with $\mathcal{V}'_t$.
The resulting representation is processed by the same visual encoder and projection layers as used for video encoding.
Finally, the multimodal embeddings corresponding to $<$$\mathcal{A}, \mathcal{V}, \mathcal{R}, \mathcal{V}_t, \mathcal{M}_t, \mathcal{V}'_t$$>$ are injected into a predefined system prompt (Sec.~\ref{sec:prompt_system_instrution}) and fed into the LLM backbone, LLaMA-2-7B-Chat~\cite{touvron2023llama}.
As shown in Fig.~\ref{fig:auditor_model}, the LLM output includes natural-language quality analysis, IoU estimation, and predictions of the mask type and recommended action (additional instruction-tuning prompts are provided in Sec.~\ref{sec:prompt_llm_instruction}).

\noindent\textbf{Training.}
The pretraining stage for audio-/visual-text alignment is conducted following~\cite{zhou2025think}.
Subsequently, we perform instruction tuning using parameter-efficient LoRA layers, with a rank of 32 and a scaling factor of 64.
The batch size is set to 4.
The training set consists of 1,306 videos, yielding 16,761 $<$video, reference, mask$>$ samples (Table~\ref{tab:MQ-RAVSBench}), among which \textit{perfect} masks constitute only a small fraction.
To balance positive and negative samples, we adopt a sampling strategy in which each epoch contains 1,306 samples, with one mask selected per video, and enforce two \textit{perfect} masks per mini-batch (positive sample ratio $p=50\%$).
This strategy allows the model to gradually observe all 16,761 samples across multiple epochs.
In practice, we train MQ-Auditor for 48 epochs on four NVIDIA A100 GPUs (40GB) using bf16 precision. %, which takes approximately 11 hours.
We use AdamW optimizer with an initial learning rate of $1 \times 10^{-4}$.

\noindent\textbf{Evaluation.}
For the \textit{image-based} evaluation protocol, MQ-Auditor processes each video using a single key frame and its associated candidate mask.
The RMSE and ${F}_2$-score are computed by averaging results over all frame-level samples (Eqs.~\ref{eq:RMSE_for_image_based}\&\ref{eq:Fscore_for_image_based}).
For the \textit{video-based} evaluation protocol, MQ-Auditor simultaneously considers predictions of all video frames.
Frame-level predictions are first summarized to obtain video-level RMSE and ${F}_2$-scores, which are then averaged across all test videos (Eqs.~\ref{eq:RMSE_for_video_based}\&\ref{eq:Fscore_for_video_based}).
The training and evaluation codes will be released to facilitate reproducibility.

\begin{table*}[t]
\centering
\caption{Video-based evaluation of MQ-Auditor on MQ-RAVSBench.
For the {Merge} and {Full\_neg} types, we primarily evaluate the hard samples with the highest mask IoU.
$F_2$-M and $F_2$-A denote the $F_2$-scores for mask type and action predictions, respectively.}

\vspace{-1.5ex}
\label{tab:video_based_evaluation_main}
\scriptsize
\setlength{\tabcolsep}{8pt} % 稍微增加列间距，让整体均匀分布
\resizebox{1.0\textwidth}{!}{
% \renewcommand{\arraystretch}{0.8} % 将行高缩减为原来的 80%
% \begin{tabularx}{\textwidth}{ll c cc cc cc c c c}
\begin{tabular}{ll c cc cc cc c c c}
\toprule
\multirow{2.5}{*}{\textbf{Settings}} & \multirow{2.5}{*}{\textbf{Metrics}} & \textbf{Perfect} & \multicolumn{2}{c}{\textbf{Cutout}} & \multicolumn{2}{c}{\textbf{Dilate}} & \multicolumn{2}{c}{\textbf{Erode}} & \textbf{Merge} & \textbf{Full\_neg} & \multirow{2.5}{*}{\textbf{Avg.}}\\ 
\cmidrule(lr){3-3} \cmidrule(lr){4-5} \cmidrule(lr){6-7} \cmidrule(lr){8-9} \cmidrule(lr){10-10} \cmidrule(lr){11-11}
 &  & All & H & M & H & M & H & M & H & H & \\ % 这里最后一列留空，不放横杠
\midrule

% Seen Setting
\multirow{3}{*}{{Seen}} 
 & RMSE $\downarrow$  & 0.276 & 0.163 & 0.148 & 0.231 & 0.180 & 0.279 & 0.251 & 0.243 & 0.556 & 0.258 \\
 & $F_2$-M $\uparrow$ & 57.02 & 86.17 & 93.24 & 61.68 & 75.64 & 29.78 & 42.85 & 51.16 & 55.77 & 61.48  \\
 & $F_2$-A $\uparrow$ & 57.02 & 34.92 & 84.57 & 21.24 & 70.76 & 11.76 & 40.02 & 40.99 & 58.34 & 46.63\\ 

\midrule % 如果有第二组数据，这里加一条线区分

% 示例：如果是 Unseen
\multirow{3}{*}{{Unseen}} 
 & RMSE $\downarrow$  & 0.185 & 0.199 & 0.183 & 0.190 & 0.170 & 0.190 & 0.222 & 0.299 & 0.567 & 0.245 \\
 & $F_2$-M $\uparrow$ & 67.72 & 87.26 & 94.27 & 57.86 & 82.62 & 28.35 & 39.44 & 53.83 & 49.47 & 62.31  \\
 & $F_2$-A $\uparrow$ & 67.72 & 45.83 & 76.91 & 25.31 & 70.04 & 26.30 & 29.08 & 45.61 & 52.37 & 48.80 \\ 

\bottomrule
\end{tabular}
% \renewcommand{\arraystretch}{1} % 恢复默认（如果在环境外修改）
% \end{tabularx}
}
\end{table*}

\begin{table*}[t]
\centering
\vspace{-1ex}
\caption{Ablation study on the utilization of mask information. For Cutout, Dilate, and Erode mask types, we first evaluate hard (H) and medium-hard (M) samples separately and report their averaged results. {`Avg.'} denotes the mean value across all columns.}
\label{tab:ab_on_mask_utilization}
\vspace{-1.5ex}
\small
\setlength{\tabcolsep}{5pt} % 稍微减小列间距以适应新列
\resizebox{1.0\textwidth}{!}{
\begin{tabular}{ll|cccccc|cccccc|c} % 增加了一个 l 列
\toprule
\multirow{2.5}{*}{\textbf{Metrics}} & \multirow{2.5}{*}{\textbf{Strategies}} & \multicolumn{6}{c|}{\textbf{Seen}} & \multicolumn{6}{c|}{\textbf{Unseen}} & \multirow{2.5}{*}{\textbf{Avg.}} \\
\cmidrule(lr){3-8} \cmidrule(lr){9-14}
 & & Perfect & Cutout & Dilate & Erode & Merge & Full\_neg & Perfect & Cutout & Dilate & Erode & Merge & Full\_neg & \\
\midrule

% 使用 multirow 合并三行
\multirow{3}{*}{RMSE $\downarrow$} 
 & Mask Only ($\mathcal{M}_t$) & \textbf{0.294} & 0.186 & \textbf{0.248} & \textbf{0.287} & 0.357 & 0.588 & 0.304 & 0.176 & 0.238 & 0.306 & 0.364 & 0.604 & 0.329 \\
 & Masked Frame ($\mathcal{V}'_t$) & 0.523 & 0.304 & 0.341 & 0.453 & 0.367 & \textbf{0.383} & 0.431 & 0.249 & 0.300 & 0.400 & 0.351 & \textbf{0.407} & 0.376 \\
 & \textbf{Both ($[\mathcal{M}_t;\mathcal{V}'_t]$)} & 0.359 & \textbf{0.180} & 0.265 & 0.317 & \textbf{0.334} & 0.507 & \textbf{0.274} & \textbf{0.148} & \textbf{0.179} & \textbf{0.263} & \textbf{0.310} & 0.508 & \textbf{0.303} \\ \midrule

% 使用 multirow 合并三行
\multirow{3}{*}{$F_2$-M $\uparrow$} 
 & Mask Only ($\mathcal{M}_t$) & \textbf{75.32} & 86.25 & 38.44 & 32.57 & 59.63 & 63.84 & \textbf{77.41} & \textbf{95.32} & 63.00 & 24.46 & 59.98 & 63.28 & 61.62 \\
 & Masked Frame ($\mathcal{V}'_t$) & 56.12 & 71.88 & 59.47 & 20.40 & 72.55 & \textbf{81.78} & 54.74 & 78.98 & 72.72 & 15.39 & 76.02 & \textbf{81.78} & 61.82 \\
 & \textbf{Both ($[\mathcal{M}_t;\mathcal{V}'_t]$)} & 60.58 & \textbf{87.60} & \textbf{69.15} & \textbf{41.62} & \textbf{74.55} & 69.43 & 64.34 & 95.18 & \textbf{82.83} & \textbf{34.28} & \textbf{77.73} & 70.92 & \textbf{69.02} \\
 \midrule

% 使用 multirow 合并三行
\multirow{3}{*}{$F_2$-A $\uparrow$} 
 & Mask Only ($\mathcal{M}_t$) & \textbf{75.32} & 56.46 & 28.13 & 26.65 & 36.36 & 67.46 & \textbf{77.41} & \textbf{64.10} & 38.73 & 20.33 & 33.39 & 66.30 & 49.22 \\
 & Masked Frame ($\mathcal{V}'_t$) & 56.12 & 48.47 & 42.85 & 18.71 & 39.11 & \textbf{85.51} & 54.74 & 51.81 & 49.20 & 21.69 & 39.27 & \textbf{84.33} & 49.32 \\
 & \textbf{Both ($[\mathcal{M}_t;\mathcal{V}'_t]$)} & 60.58 & \textbf{57.57} & \textbf{48.16} & \textbf{32.06} & \textbf{43.00} & 74.52 & 64.34 & 62.29 & \textbf{59.13} & \textbf{29.68} & \textbf{44.57} & 74.41 & \textbf{54.19} \\

\bottomrule
\end{tabular}
}
% \vspace{-5ex}
\end{table*}

\begin{table*}[htp]
\centering
% \vspace{-1ex}
\caption{Ablation study on IoU estimation strategies.
`Direct NTP' denotes generating the IoU value via direct next-token prediction.
`Regress ST' denotes first generating a special token (\textit{e.g.}, $<$iou\_value$>$), followed by regressing the IoU using a linear layer.}

\label{tab:ab_iou_estimation}
\vspace{-1.5ex}
\small
\setlength{\tabcolsep}{5pt} % 稍微减小列间距以适应新列
\resizebox{1.0\textwidth}{!}{
\begin{tabular}{ll|cccccc|cccccc|c} % 增加了一个 l 列
\toprule
\multirow{2.5}{*}{\textbf{Metrics}} & \multirow{2.5}{*}{\textbf{Strategies}} & \multicolumn{6}{c|}{\textbf{Seen}} & \multicolumn{6}{c|}{\textbf{Unseen}} & \multirow{2.5}{*}{\textbf{Avg.}} \\
\cmidrule(lr){3-8} \cmidrule(lr){9-14}
 & & Perfect & Cutout & Dilate & Erode & Merge & Full\_neg & Perfect & Cutout & Dilate & Erode & Merge & Full\_neg & \\
\midrule

% 使用 multirow 合并三行
\multirow{2}{*}{RMSE $\downarrow$} 
 & Regress ST & 0.758 & 0.680 & 0.590 & 0.612 & 0.474 & \textbf{0.226} & 0.565 & 0.618 & 0.459 & 0.471 & 0.412 & \textbf{0.298} & 0.514 \\
 & Direct NTP & \textbf{0.359} & \textbf{0.180} & \textbf{0.265} & \textbf{0.317} & \textbf{0.334} & 0.507 & \textbf{0.274} & \textbf{0.148} & \textbf{0.179} & \textbf{0.263} & \textbf{0.310} & 0.508 & \textbf{0.303} \\ \midrule

% 使用 multirow 合并三行
\multirow{2}{*}{$F_2$-M $\uparrow$} 
 & Regress ST  & \textbf{69.03} & 85.83 & 58.75 & 34.52 & 60.46 & 68.77 & \textbf{74.44} & 93.95 & 78.00 & \textbf{36.04} & 58.49 & \textbf{71.51} & 65.82 \\
 & Direct NTP & 60.58 & \textbf{87.60} & \textbf{69.15} & \textbf{41.62} & \textbf{74.55} & \textbf{69.43} & 64.34 & \textbf{95.18} & \textbf{82.83} & 34.28 & \textbf{77.73} & 70.92 & \textbf{69.02} \\
 \midrule

% % 使用 multirow 合并三行
\multirow{2}{*}{$F_2$-A $\uparrow$} 
 & Regress ST  & \textbf{69.03} & 54.09 & 40.32 & 27.16 & 40.34 & 70.66 & \textbf{74.44} & 55.99 & 50.99 & 28.78 & 36.72 & 72.56 & 51.76 \\
 & Direct NTP & 60.58 & \textbf{57.57} & \textbf{48.16} & \textbf{32.06} & \textbf{43.00} & \textbf{74.52} & 64.34 & \textbf{62.29} & \textbf{59.13} & \textbf{29.68} & \textbf{44.57} & \textbf{74.41} & \textbf{54.19} \\

\bottomrule
\end{tabular}
\vspace{-6ex}
}
\end{table*}

\begin{table*}[t]
\centering
\vspace{-1.5ex}
\caption{Ablation study on the per-batch positive sample ratio $p$ in model training.}
\label{tab:ab_positive_sample_ration}
\vspace{-1.5ex}
\small
\setlength{\tabcolsep}{5pt} % 稍微减小列间距以适应新列
\resizebox{1.0\textwidth}{!}{
\begin{tabular}{lc|cccccc|cccccc|c} % 增加了一个 l 列
\toprule
\multirow{2.5}{*}{\textbf{Metrics}} & \multirow{2.5}{*}{\textbf{Ratio $p$}} & \multicolumn{6}{c|}{\textbf{Seen}} & \multicolumn{6}{c|}{\textbf{Unseen}} & \multirow{2.5}{*}{\textbf{Avg.}} \\
\cmidrule(lr){3-8} \cmidrule(lr){9-14}
 & & Perfect & Cutout & Dilate & Erode & Merge & Full\_neg & Perfect & Cutout & Dilate & Erode & Merge & Full\_neg & \\
\midrule

% 使用 multirow 合并三行
\multirow{3}{*}{RMSE $\downarrow$} 
 & 25\% &  0.409 & 0.200 & 0.280 & 0.348 & 0.353 & \textbf{0.420} & 0.378 & 0.173 & 0.270 & 0.304 & 0.345 & \textbf{0.440} & 0.327 \\
 & 50\% & 0.359 & \textbf{0.180} & \textbf{0.265} & 0.317 & \textbf{0.334} & 0.507 & 0.274 & \textbf{0.148} & \textbf{0.179} & 0.263 & \textbf{0.310} & 0.508 & \textbf{0.303} \\ 
 & 75\% & \textbf{0.268} & 0.207 & 0.286 & \textbf{0.310} & 0.362 & 0.652 & \textbf{0.156} & 0.168 & 0.187 & \textbf{0.239} & 0.368 & 0.716 & 0.327 \\ \midrule

% 使用 multirow 合并三行
\multirow{3}{*}{$F_2$-M $\uparrow$} 
 & 25\% & 40.24 & \textbf{88.47} & \textbf{77.23} & \textbf{51.69} & \textbf{77.94} & \textbf{76.99} & 52.77 & 94.91 & \textbf{83.21} & \textbf{46.30} & \textbf{78.52} & \textbf{77.05} & \textbf{70.44} \\
 & 50\% & 60.58 & 87.60 & 69.15 & 41.62 & 74.55 & 69.43 & 64.34 & \textbf{95.18} & 82.83 & 34.28 & 77.73 & 70.92 & 69.02 \\ 
 & 75\% & \textbf{86.37} & 81.43 & 36.89 & 12.40 & 60.31 & 55.17 & \textbf{89.99} & 92.17 & 53.60 & 14.01 & 48.85 & 49.50 & 56.72 \\
 \midrule

% 使用 multirow 合并三行
\multirow{3}{*}{$F_2$-A $\uparrow$} 
 & 25\% & 40.24 & \textbf{60.32} & \textbf{53.17} & \textbf{39.71} & \textbf{44.51} & \textbf{82.26} & 52.77 & \textbf{66.00} & 54.78 & \textbf{37.95} & 44.15 & \textbf{80.55} & \textbf{54.70} \\
 & 50\% & 60.58 & 57.57 & 48.16 & 32.06 & 43.00 & 74.52 & 64.34 & 62.29 & \textbf{59.13} & 29.68 & \textbf{44.57} & 74.41 & 54.19 \\
 & 75\% & \textbf{86.37} & 52.18 & 26.36 & 10.98 & 32.18 & 59.62 & \textbf{89.99} & 59.77 & 35.62 & 16.00 & 27.75 & 50.92 & 45.64 \\

\bottomrule
\end{tabular}
}
\vspace{-3ex}
\end{table*}

\begin{table}[t] % 去掉星号，变为单栏
\centering
% \caption{Performance comparison of prior segmentation methods (EEMC~\cite{wang2024ref} and TGS-Agent~\cite{zhou2025think}) before and after using our MQ-Auditor for assessment. }
\caption{Performance comparison of prior Ref-AVS segmentation models (EEMC~\cite{wang2024ref} and TGS-Agent~\cite{zhou2025think}) before and after applying MQ-Auditor for mask quality assessment.}

\vspace{-1.5ex}
\label{tab:exp_before_and_after_auditor}
\small
% 减小列间距以适应单栏宽度 (根据实际效果可微调 3pt-5pt)
% \setlength{\tabcolsep}{5pt} 
\resizebox{1.0\columnwidth}{!}{

\begin{tabular}{l ccc ccc}
\toprule
\multirow{2}{*}{\textbf{Methods}} & \multicolumn{3}{c}{\textbf{Seen}} & \multicolumn{3}{c}{\textbf{Unseen}} \\
\cmidrule(lr){2-4} \cmidrule(lr){5-7}
 & $\mathcal{J}$ & $\mathcal{F}$ & $\mathcal{J}$\&$\mathcal{F}$ & $\mathcal{J}$ & $\mathcal{F}$ & $\mathcal{J}$\&$\mathcal{F}$ \\
\midrule

EEMC            & 31.9 & 42.2 & 37.1 & 16.7 & 32.5 & 24.6 \\
\textbf{+ MQ-Auditor} & \textbf{38.4} & \textbf{50.6} & \textbf{44.5} & \textbf{56.7} & \textbf{66.0} & \textbf{61.3}\\
\midrule

TGS-Agent       & 36.4 & 47.6 & 42.0 & 44.9 & 53.8 & 49.3 \\
\textbf{+ MQ-Auditor} & \textbf{37.1} & \textbf{49.6} & \textbf{43.4} & \textbf{46.7} & \textbf{55.7} & \textbf{51.2} \\

\bottomrule
\end{tabular}
}
% \vspace{-6ex}
\end{table}

\section{Experiments}

% We evaluate our method on the proposed MQ-RAVSBench using the introduced evaluation strategies (Sec.~\ref{sec:model}) and metrics (Sec.~\ref{sec:evaluation_details}).

\subsection{Main Results}\label{sec:exp_main_results}
We benchmark MQ-RAVSBench by comparing MQ-Auditor with several state-of-the-art open-source and closed-source MLLMs, including Video-LLaMA3-7B~\cite{zhang2025videollama3}, Qwen2.5-Omni-7B~\cite{xu2025qwen25omni}, Ming-Flash-Omni~\cite{ai2025mingomni}, and Gemini-3-Flash-Preview~\cite{google2025gemini3}.
Notably, Ming-Flash-Omni adopts a sparse Mixture-of-Experts (MoE) architecture with 100B total parameters, of which only 6.1B are activated per token.
All compared models are capable of directly processing and reasoning over both audio and video inputs, making the comparison fair and informative.
Unless otherwise specified, we conduct the comparison using the image-based evaluation protocol.
The quantitative results are summarized in Table~\ref{tab:image_based_evaluation_main}.
We observe that Video-LLaMA3 and Qwen2.5-Omni tend to accept most candidate masks, leading to strong performance on the \textit{Perfect} mask type but substantially weaker results on negative mask types.
In contrast, Ming-Flash-Omni exhibits an overly conservative behavior, frequently rejecting candidate masks, which results in poor performance on \textit{Perfect} masks but relatively stronger results on the \textit{Full\_neg} type.
The more powerful Gemini-3-Flash demonstrates comparatively balanced performance across different mask types and is particularly effective at identifying \textit{Full\_neg} masks.
Nevertheless, MQ-Auditor consistently outperforms all competing models across the majority of mask types in both Seen and Unseen settings, achieving the best overall average performance ({`Avg.'}).
\textit{\textbf{Qualitative comparisons are provided in Sec.~\ref{sec:qualitative_analysis} (Figs.~\ref{fig:audit_comparison_sample_1}$\sim$\ref{fig:audit_comparison_sample_6}).}}
MQ-Auditor is also inference-efficient (see Table~\ref{tab:efficiency_comparison}).
These results indicate that although existing MLLMs demonstrate strong capabilities in general audio-visual understanding, such as captioning and question answering, they remain inadequate for the specialized and fine-grained MQA-RefAVS task.
Table~\ref{tab:video_based_evaluation_main} further reports MQ-Auditor's performance under the video-based evaluation protocol, completing our benchmark analysis.

\subsection{Ablation Studies}\label{sec:ablations}
In this section, we report ablation study results under the image-based evaluation protocol. 

\noindent\textbf{Effect of Mask Utilization.}
Our method first applies the candidate mask $\mathcal{M}_t$ to raw video frame $\mathcal{V}_t$ to obtain a masked frame $\mathcal{V}'_t$, which is then concatenated with the original mask $\mathcal{M}_t$ to extract input embeddings for MQ-Auditor.
We study the impact of mask usage by considering two variants: using only the mask $\mathcal{M}_t$ or only the masked frame $\mathcal{V}'_t$.
As shown in Table~\ref{tab:ab_on_mask_utilization}, using the mask alone benefits the assessment of \textit{Perfect} and \textit{Cutout} types, as the binary mask provides direct \textit{geometric} cues.
In contrast, using only the masked frame is particularly effective for the \textit{Full\_neg} type, since the masked frame clearly reveals the \textit{semantic} content captured by the mask.
Combining both representations allows the model to leverage complementary geometric and semantic information, resulting in best overall performance.

\noindent\textbf{Different IoU Estimation Manners.}
MQ-Auditor directly estimates IoU by generating the IoU value in a next-token prediction manner, denoted as {`Direct NTP'}.
Alternatively, IoU can be treated as a single scalar represented by a special token (\textit{e.g.}, $<$iou\_value$>$).
In this case, MQ-Auditor first generates the special token, and the corresponding embedding from final LLM decoder layer is fed into a linear regression head with a sigmoid activation, denoted as {`Regress ST'}.
The mean squared error between the regressed IoU and the ground truth is combined with the next-token prediction loss for training.
As shown in Table~\ref{tab:ab_iou_estimation}, the {`Regress ST'} strategy is particularly effective for the \textit{Full\_neg} mask type (IoU = 0), yielding a substantially lower RMSE.
However, its performance on other mask types is notably worse than that of {`Direct NTP'}.
This may because IoU values for these mask types span a wide range in $(0,1)$ and can be difficult to accurately regress using a single linear layer.

\noindent\textbf{Training Positive Sample Ratio $p$.}
As discussed in Sec.~\ref{sec:model} (``Training''), we define the positive sample ratio $p$ as the proportion of positive samples (\textit{i.e.}, instances with \textit{perfect} masks) within each mini-batch.
Table~\ref{tab:ab_positive_sample_ration} presents ablation results for different values of $p$.
A higher ratio ($p=75\%$) reduces RMSE and improves $F_2$ scores for the \textit{Perfect} type, but leads to a substantial performance drop on negative mask types.
Conversely, when $p=25\%$, the model achieves higher $F_2$ scores across most negative mask types, while RMSE remains relatively high.
Overall, $p=50\%$ provides the best trade-off between IoU estimation accuracy and classification performance across all mask types.

% \subsection{Practical Use of MQ-Auditor}
\subsection{Segmentation Improvement via MQ-Auditor}
Results in Sec.~\ref{sec:exp_main_results} demonstrate the effectiveness and superiority of MQ-Auditor in assessing candidate mask quality.
Notably, the MQ-Auditor does not require access to ground-truth masks to estimate IoU or predict mask types and recommended actions during inference.
This property allows MQ-Auditor to be seamlessly integrated with existing Ref-AVS segmentation models in practical scenarios.
Specifically, we validate this capability on two prior SOTA methods, EEMC~\cite{wang2024ref} and TGS-Agent~\cite{zhou2025think}.
These methods are first used to generate segmentation masks, which are then assessed by MQ-Auditor.
We identify video samples whose masks are predicted as the \textit{Full\_neg} type by MQ-Auditor, which represents the most frequent failure mode, where the segmentation model incorrectly segments a non-target object.
Since MQ-Auditor also produces the target object category in its reasoning output, we leverage this information to prompt Grounded-SAM2~\cite{ren2024grounded} to re-generate the segmentation masks.
We report segmentation performance before and after applying MQ-Auditor-based quality assessment.
As shown in Table~\ref{tab:exp_before_and_after_auditor}, MQ-Auditor consistently improves the performance of both baseline methods.
In particular, for the EEMC baseline, the Jaccard index $\mathcal{J}$ and F-score $\mathcal{F}$ are improved by 40\% and 33.5\% for Unseen test samples, respectively.
In this setting, MQ-Auditor acts as a \textit{reflection} agent that automatically diagnoses mask quality and provides actionable guidance, including corrected target object cues, to support mask revision.
\textbf{\textit{Qualitative examples are shown in Figs.~\ref{fig:eemc_before_after_audit_sample_1} and \ref{fig:eemc_before_after_audit_sample_2}}}.
We believe that such improvements can be further amplified as more powerful mask quality assessment models emerge in the future.

\section{Conclusion}
In this work, we study an overlooked problem, Mask Quality Assessment (MQA), within language-referred audio-visual segmentation, which requires assessing segmentation masks without access to ground-truth annotations at test time.
We construct the MQ-RAVSBench dataset and propose a baseline approach, MQ-Auditor.
Extensive experiments show that MQ-Auditor outperforms strong open-source and commercial MLLMs and can effectively complement existing Ref-AVS systems.
We hope this work encourages future research to move beyond mask generation alone and toward segmentation systems equipped with explicit error diagnosis, quality auditing, and automatic correction mechanisms.

% \clearpage
\noindent\textbf{Broader Impact.}
This work aims to improve the reliability and interpretability of multimodal segmentation systems by enabling automatic quality assessment of segmentation masks.
By providing structured and actionable feedback, the proposed MQA framework, MQ-Auditor, can help reduce silent failures in downstream segmentation applications.
However, as with other MLLM-based models, MQ-Auditor may inherit biases from its training data and underlying large language models, which could affect the consistency of error assessment across different object categories or environments.
Therefore, caution is required when deploying such systems in safety-critical scenarios, where human oversight remains essential.

% Overall, we believe the benefits of improved transparency and robustness outweigh the risks, and this work contributes positively to the development of trustworthy machine learning systems.


%%%%%%%%% REFERENCES
{\small
\bibliographystyle{ieee_fullname}
\bibliography{egbib}
}

%%%%%%%%%%%%%%%%%%%%%%%%%%%%%%%%%%%%%%%%%%%%%%%%%%%%%%%%%%%%%%%%%%%%%%%%%%%%%%%
%%%%%%%%%%%%%%%%%%%%%%%%%%%%%%%%%%%%%%%%%%%%%%%%%%%%%%%%%%%%%%%%%%%%%%%%%%%%%%%
% APPENDIX
%%%%%%%%%%%%%%%%%%%%%%%%%%%%%%%%%%%%%%%%%%%%%%%%%%%%%%%%%%%%%%%%%%%%%%%%%%%%%%%
%%%%%%%%%%%%%%%%%%%%%%%%%%%%%%%%%%%%%%%%%%%%%%%%%%%%%%%%%%%%%%%%%%%%%%%%%%%%%%%
\newpage
\appendix
\onecolumn
\clearpage

\section{Related Work}
\noindent\textbf{Audio-Visual Scene Understanding}
aims to analyze audio and visual signals, particularly their correspondence and complementarity, to perceive and reason about dynamic audio-visual scenes~\cite{li2024object,li2025patch,shen2023fine,mao2024tavgbench,zhou2025mettle,kryklyvets2025mavis,kurpath2025benchmark,zhou2026mtavgbench}.
Early research in this area primarily focused on \textit{temporal} understanding.
For example, audio-visual event localization~\cite{tian2018audio,lin2019dual,xia2022cross,rao2022dual,zhou2021psp,zhou2022cpsp,liu2025towards,gao2025learning} and audio-visual video parsing~\cite{tian2020unified,wu2021exploring,cheng2022joint,lai2023modality,zhou2024advancing,zhou2024label,zhao2025multimodal} aim to identify audio and visual events along the temporal axis.
Subsequent works extend these settings to more challenging scenarios, including open-vocabulary~\cite{zhou2025towards}, online~\cite{yu2025prefm}, portrait-mode~\cite{liu2025audio}, and untrimmed videos~\cite{geng2023dense,zhou2025dense,zhou2025clasp}.
More recently, research attention has shifted toward finer-grained \textit{spatial} (and spatio-temporal) understanding.
A representative direction is audio-visual segmentation, which aims to localize sounding objects in video frames at the object level~\cite{zhou2022avs,li2023catr,mao2023multimodal,gao2024avsegformer,ma2024stepping}, semantic level~\cite{zhou2023avss,guo2024enhance,zhoualoha,luo2025tavis,liu2025dynamic,huang2025revisiting}, or instance level~\cite{guo2025audio,seo2025learning}.
Similar to these tasks, Ref-AVS also requires spatial-temporal understanding of audio-visual scenes, and we review the most relevant works below.

\vspace{4ex}
\noindent\textbf{Referring Audio-Visual Segmentation}
aims to segment target objects in audible videos according to given referring expressions.
Early efforts~\cite{wang2024ref,radman2025tsam,wang2025sam2love} typically adopt multimodal fusion strategies that integrate cross-modal cues before prompting a segmentation decoder such as Mask2Former~\cite{cheng2021mask2former}, SAM~\cite{kirillov2023sam}, or SAM2~\cite{ravi2024sam2}.
For instance, SAM2-LOVE~\cite{wang2025sam2love} proposes dedicated token propagation and accumulation mechanisms to compress multimodal cues across video frames into a single \texttt{[SEG]} token, which is then used to prompt SAM2 for segmentation.
More recently, several approaches leverage powerful multimodal large language models (MLLMs) to generate fused \texttt{[SEG]} tokens~\cite{du2025crab,luo2025aurora,omniavs,jin2025simtoken}, followed by a mask decoder.
For example, AURORA~\cite{luo2025aurora} employs Video-LLaMA2~\cite{cheng2024videollama2}, while OmniAVS~\cite{omniavs} adopts InternVL2~\cite{chen2024internvl} augmented with an additional audio encoder.
Instead of relying on \texttt{[SEG]} tokens, TGS-Agent~\cite{zhou2025think} directly generates explicit textual descriptions of the target object to guide subsequent detection and segmentation.
These prior works primarily focus on improving \textit{segmentation mask generation}, whereas our work addresses an orthogonal problem of \textit{mask quality assessment}, which requires not only understanding the Ref-AVS task but also explicitly reasoning about the correctness and reliability of generated masks. Moreover, effective mask quality assessment can further improve segmentation mask generation.

\vspace{4ex}
\noindent\textbf{Image \& Video Quality Assessment}
aim to automatically predict human perceptual judgments of visual content, typically expressed as mean opinion scores (MOS).
Image quality assessment (IQA) focuses on static images affected by artifacts such as \textit{blur}, \textit{noise}, and \textit{color distortion}, while video quality assessment (VQA) extends this problem to videos, where temporal artifacts including \textit{motion inconsistency}, \textit{frame drops}, and \textit{temporal flicker} must also be considered.
To address IQA, early studies directly regress MOS using transformer-based backbones~\cite{ke2021musiq,yang2024align,xu2024boosting} or MLLMs~\cite{you2025teaching,wu2023qalign}.
More recent methods increasingly formulate IQA as an instruction-following and reasoning task rather than a pure score-regression problem.
For example, Q-Instruct~\cite{wu2024qinstruct} and DepictQA~\cite{you2024depicting} construct large-scale datasets with natural-language feedback on low-level image quality attributes and demonstrate that targeted instruction tuning can transform general-purpose MLLMs into effective low-level quality experts.
Subsequently, Q-Insight~\cite{li2025qinsight} and Q-Ponder~\cite{cai2025qponder} employ reinforcement learning to further improve cross-domain score accuracy and natural-language quality reasoning.
In the VQA domain, Q-Bench-Video~\cite{zhang2025q} and FineVQ~\cite{duan2025finevq} benchmark the video-quality understanding capabilities of MLLMs, revealing that generic models still struggle with temporal artifacts and subtle degradations.
LMM-VQA~\cite{ge2025lmm} adapts large vision--language models to video by introducing video-specific spatio-temporal tokenization strategies.
VQAThinker~\cite{cao2025vqathinker} extends reinforcement-learning-based reasoning to VQA by incorporating multiple targeted reward functions, improving both generalization and explainability.
Unlike IQA and VQA, which assess \textit{perceptual} or \textit{aesthetic} quality, the studied MQA focuses on evaluating the \textit{semantic fidelity} of segmentation masks with respect to multimodal referring inputs.
\clearpage

% \vspace*{\fill}

\begin{table*}[hp]
\centering
\tiny
\caption{Number of samples for each mask type in MQ-RAVSBench.
For each $<$video, reference$>$ pair, we generate 1, 2, and up to 3 masks for the \textit{Perfect}, \textit{Cutout/Dilate/Erode}, and \textit{Merge/Full\_neg} types, respectively.
For the \textit{image-based} evaluation, mask generation is performed on a single key video frame, whereas for the \textit{video-based} evaluation, masks are generated for all 10 video frames.}
\resizebox{1.0\textwidth}{!}{
% \begin{tabular}{l S[table-format=4.0] S[table-format=4.0] S[table-format=5.0]}
\begin{tabular}{l|r|rrrrrr}
\toprule
 & \textbf{Total} & \textbf{\#Perfect} & \textbf{\#Cutout} &\textbf{\#Dilate} &\textbf{\#Erode} & \textbf{\#Merge} & \textbf{\#Full\_neg} \\ \midrule
Train &  16,761 & 1,306 & 2,612 & 2,612 & 2,612 & 3,809 & 3,810 \\ \midrule
\textit{\textbf{image-based evaluation}} & & & & & & & \\ 
Test (Seen) & 5,609 & 437 & 874 & 874 & 874 & 1,274 & 1,276 \\
Test (Unseen) & 3,691 & 303 & 606 & 606 & 606 & 785 & 785 \\
% 使用 \multicolumn 保持右对齐，并确保与上方列对齐
% \multicolumn{1}{r}{total} & 9,300 &  \\ \midrule
\textit{\textbf{video-based evaluation}} & & & & & & & \\ 
Test (Seen) & 7,674 & 600 & 1,200 & 1,200 & 1,200 & 1,738 & 1,736 \\
Test (Unseen) &  4,966 & 400 & 800 & 800 & 800 & 1,081 & 1,085 \\
% \multicolumn{1}{r}{total} & 100 & 100 & 12,640 \\ 
\bottomrule
\end{tabular}
}
\label{tab:data_stat_of_each_mask_type}
\end{table*}

\begin{figure*}[hp]
  \centering
\includegraphics[width=\textwidth]{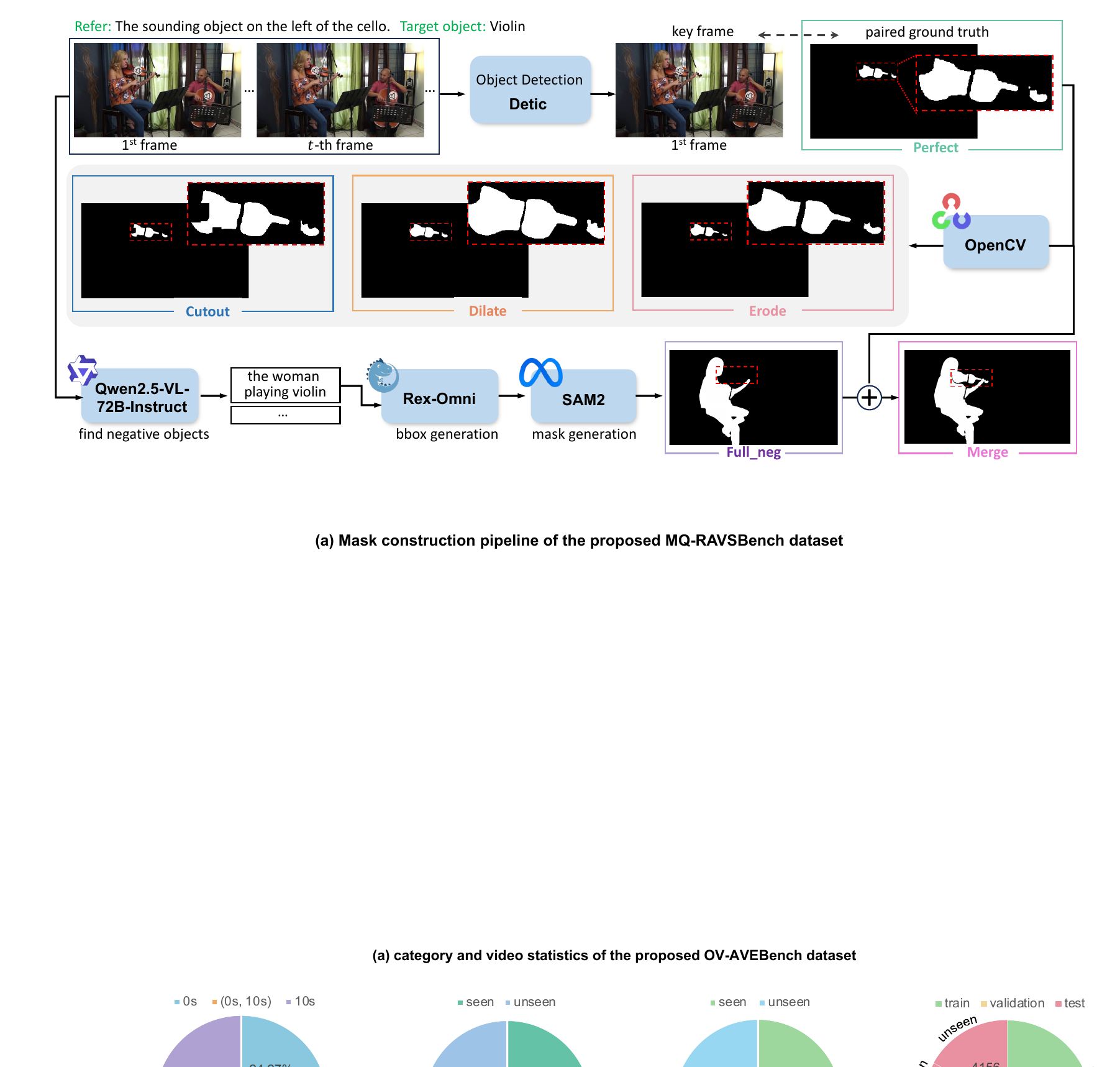}
   
   \caption{IoU distribution of MQ-RAVSBench.
    For the test set, IoU statistics are computed based on samples used in the \textit{image-based} evaluation.
    The IoU values for the \textit{Perfect} and \textit{Full\_neg} types are always 1 and 0, respectively.
    The \textit{Cutout/Dilate/Erode} masks typically exhibit higher IoU values around 0.8; we intentionally control this range to avoid overly obvious quality errors that would trivialize assessment.
    The IoU values of \textit{Merge} masks span the full range from 0 to 1, depending on the relative area between the ground-truth object and the merged negative regions.
    For example, when the ground-truth object is small and the merged negative objects are large, the resulting \textit{Merge} mask yields a low IoU; otherwise, a higher IoU is obtained.}
   \label{fig:iou_distribution}
\end{figure*}
% \vspace*{\fill}

\section{More Dataset Statistics}\label{sec:appendix_more_data_stat}
Table~\ref{tab:MQ-RAVSBench} in main paper summarizes the overall statistics of videos and mask samples in MQ-RAVSBench.
In Table~\ref{tab:data_stat_of_each_mask_type}, we further reports the detailed breakdown of samples for each mask type.
In addition, in Fig.~\ref{fig:iou_distribution}, we visualize the IoU distribution of samples used in the \textit{image-based} evaluation.
Additional details can be found in the corresponding table/figure captions.

\section{Calculation Details of Evaluation Metrics}\label{sec:supp_metric_details}
We provide detailed formulations for computing the evaluation metrics, \textit{i.e.}, RMSE and the $F_2$-score.

\textbf{RMSE.}
This metric computes the Root Mean Square Error (RMSE) between the predicted IoU $s$ and the ground-truth IoU $s_g$ over all $N$ evaluation samples:
\begin{equation}
    \text{RMSE} = \sqrt{ \frac{1}{N} \sum_{i=1}^{N} (s - s_g)^2 }.
    \label{eq:RMSE}
\end{equation}

$\bm{F_2}$\textbf{-score.}
Both mask type and action predictions are evaluated using the ${F_\beta}$ score with $\beta = 2$.
This choice emphasizes \textit{recall}, which is particularly desirable for MQA systems because:
\textbf{1)} missing a problematic mask is typically more costly than incorrectly flagging a correct one;
\textbf{2)} our empirical results show that MQA models tend to achieve high precision (often close to 100\%) but comparatively lower recall.
The ${F_2}$-score is computed as:
\begin{equation}
{F_{2}} ={F_{\beta}} = (1 + \beta^2) \cdot \frac{\mathrm{P} \cdot \mathrm{R}}
{\beta^2 \cdot \mathrm{P} + \mathrm{R}}, \quad 
\mathrm{P}=\frac{\mathrm{TP}}{\mathrm{TP} + \mathrm{FP}}, \quad
\mathrm{R}=\frac{\mathrm{TP}}{\mathrm{TP} + \mathrm{FN}},
\label{eq:F2_score}
\end{equation}
where $\mathrm{P}$ and $\mathrm{R}$ denote precision and recall, respectively.
$\mathrm{TP}$, $\mathrm{FP}$, and $\mathrm{FN}$ represent the numbers of true positives, false positives, and false negatives.

To be specific, our evaluation follows two strategies:\\
For the \textbf{image-based evaluation}, the final RMSE of the Seen/Unseen test set reported in Table~\ref{tab:image_based_evaluation_main} is computed by averaging over all $N_m$ image/mask samples:
\begin{equation}
    \text{RMSE} = \sqrt{ \frac{1}{N_m} \sum_{i=1}^{N_m} (s^i - s_g^i)^2 },
    \label{eq:RMSE_for_image_based}
\end{equation}
where $s^i$ and $s_g^i$ denote the predicted IoU and ground-truth IoU of the $i$-th test sample, respectively.\\
Similarly, the overall $F_2$-score is obtained by first accumulating $\mathrm{TP}$, $\mathrm{FP}$, and $\mathrm{FN}$ over all $N_m$ samples, computing per-class $F_2$, and then taking the macro average:
\begin{equation}
F_{\beta} = \frac{1}{|\mathcal{C}|} \sum_{c \in \mathcal{C}} F_{\beta}^{(c)},
\quad
F_{\beta}^{(c)} = (1+\beta^2)\cdot \frac{\mathrm{P}^{(c)}\cdot \mathrm{R}^{(c)}}{\beta^2\cdot \mathrm{P}^{(c)}+\mathrm{R}^{(c)}},
    \label{eq:Fscore_for_image_based}
\end{equation}
where $\mathcal{C}$ is the set of evaluated classes (\textit{i.e.}, mask types or actions), and $\mathrm{P}^{(c)}$, $\mathrm{R}^{(c)}$ are computed from the accumulated $\mathrm{TP}^{(c)}$, $\mathrm{FP}^{(c)}$, and $\mathrm{FN}^{(c)}$ over all samples.

For the \textbf{video-based evaluation}, the final RMSE reported in Table~\ref{tab:video_based_evaluation_main} is computed by first aggregating frame-level predictions into video-level IoU values and then averaging across all $N_v$ test videos:
\begin{equation}
    \text{RMSE} = \sqrt{ \frac{1}{N_v} \sum_{i=1}^{N_v} (\overline{s}^i - \overline{s_g}^i)^2 }, \quad
    \overline{s}^i = \frac{1}{T} \sum_{t=1}^T s^{i,t}, \quad
    \overline{s_g}^i = \frac{1}{T} \sum_{t=1}^T s_g^{i,t},
    \label{eq:RMSE_for_video_based}
\end{equation}
where $\overline{s}^i$ and $\overline{s_g}^i$ denote the predicted and ground-truth IoU values averaged over $T$ frames of the $i$-th test video, and $s^{i,t}$ and $s_g^{i,t}$ correspond to the IoU values of the $t$-th frame.

Similarly, the video-level $F_2$-score is computed by aggregating frame-level $\mathrm{TP}$, $\mathrm{FP}$, and $\mathrm{FN}$ within each video, computing per-class $F_2$, and then averaging over videos:
\begin{equation}
F_{\beta} = \frac{1}{N_v} \sum_{i=1}^{N_v}F_{\beta}^i, \quad 
F_{\beta}^i = \frac{1}{|\mathcal{C}|} \sum_{c \in \mathcal{C}} F_{\beta}^{i,(c)}.
    \label{eq:Fscore_for_video_based}
\end{equation}
Here $F_{\beta}^{i,(c)}$ is computed from the video-level aggregated $\mathrm{TP}^{i,(c)}$, $\mathrm{FP}^{i,(c)}$, and $\mathrm{FN}^{i,(c)}$ for class $c$ in the $i$-th video.

\begin{table*}[t]
\centering
\vspace{2ex}
\caption{Ablation study on the utilization of mask information. For Cutout, Dilate, and Erode mask types, we first evaluate hard (H) and medium-hard (M) samples separately and report their averaged results. {`Avg.'} denotes the mean value across all columns.}
\label{tab:supp_ab_on_mask_utilization}
% \vspace{-1.5ex}
\small
\setlength{\tabcolsep}{5pt} % 稍微减小列间距以适应新列
\resizebox{1.0\textwidth}{!}{
\begin{tabular}{ll|cccccc|cccccc|c} % 增加了一个 l 列
\toprule
\multirow{2.5}{*}{\textbf{Metrics}} & \multirow{2.5}{*}{\textbf{Strategies}} & \multicolumn{6}{c|}{\textbf{Seen}} & \multicolumn{6}{c|}{\textbf{Unseen}} & \multirow{2.5}{*}{\textbf{Avg.}} \\
\cmidrule(lr){3-8} \cmidrule(lr){9-14}
 & & Perfect & Cutout & Dilate & Erode & Merge & Full\_neg & Perfect & Cutout & Dilate & Erode & Merge & Full\_neg & \\
\midrule

% 使用 multirow 合并三行
\multirow{4}{*}{RMSE $\downarrow$} 
 & Mask Only ($\mathcal{M}_t$) & \textbf{0.294} & 0.186 & \textbf{0.248} & \textbf{0.287} & 0.357 & 0.588 & 0.304 & 0.176 & 0.238 & 0.306 & 0.364 & 0.604 & 0.329 \\
 & Masked Frame ($\mathcal{V}'_t$) & 0.523 & 0.304 & 0.341 & 0.453 & 0.367 & \textbf{0.383} & 0.431 & 0.249 & 0.300 & 0.400 & 0.351 & \textbf{0.407} & 0.376 \\
 & Mask Overlay ($\mathcal{\hat{V}}_t$) & 0.431 & 0.345 & 0.359 & 0.371 & 0.396 & 0.467 & 0.349 & 0.267 & 0.270 & 0.285 & 0.396 & 0.473 & 0.367 \\
 & \textbf{Both ($[\mathcal{M}_t;\mathcal{V}'_t]$)} & 0.359 & \textbf{0.180} & 0.265 & 0.317 & \textbf{0.334} & 0.507 & \textbf{0.274} & \textbf{0.148} & \textbf{0.179} & \textbf{0.263} & \textbf{0.310} & 0.508 & \textbf{0.303} \\ \midrule

% 使用 multirow 合并三行
\multirow{4}{*}{$F_2$-M $\uparrow$} 
 & Mask Only ($\mathcal{M}_t$) & \textbf{75.32} & 86.25 & 38.44 & 32.57 & 59.63 & 63.84 & \textbf{77.41} & \textbf{95.32} & 63.00 & 24.46 & 59.98 & 63.28 & 61.62 \\
 & Masked Frame ($\mathcal{V}'_t$) & 56.12 & 71.88 & 59.47 & 20.40 & 72.55 & \textbf{81.78} & 54.74 & 78.98 & 72.72 & 15.39 & 76.02 & \textbf{81.78} & 61.82 \\
  & Mask Overlay ($\mathcal{\hat{V}}_t$) & 71.36 & 54.38 & 36.70 & 29.27 & 47.50 & 76.29 & 76.52 & 62.90 & 64.23 & 28.99 & 48.20 & 75.10 & 55.95  \\
 & \textbf{Both ($[\mathcal{M}_t;\mathcal{V}'_t]$)} & 60.58 & \textbf{87.60} & \textbf{69.15} & \textbf{41.62} & \textbf{74.55} & 69.43 & 64.34 & 95.18 & \textbf{82.83} & \textbf{34.28} & \textbf{77.73} & 70.92 & \textbf{69.02} \\
 \midrule

% 使用 multirow 合并三行
\multirow{4}{*}{$F_2$-A $\uparrow$} 
 & Mask Only ($\mathcal{M}_t$) & \textbf{75.32} & 56.46 & 28.13 & 26.65 & 36.36 & 67.46 & \textbf{77.41} & \textbf{64.10} & 38.73 & 20.33 & 33.39 & 66.30 & 49.22 \\
 & Masked Frame ($\mathcal{V}'_t$) & 56.12 & 48.47 & 42.85 & 18.71 & 39.11 & \textbf{85.51} & 54.74 & 51.81 & 49.20 & 21.69 & 39.27 & \textbf{84.33} & 49.32 \\
  & Mask Overlay ($\mathcal{\hat{V}}_t$) & 71.36 & 39.72 & 28.75 & 29.09 & 34.83 & 79.50 & 76.52 & 40.71 & 45.52 & \textbf{39.68} & 30.63 & 78.86 & 49.60 \\
 & \textbf{Both ($[\mathcal{M}_t;\mathcal{V}'_t]$)} & 60.58 & \textbf{57.57} & \textbf{48.16} & \textbf{32.06} & \textbf{43.00} & 74.52 & 64.34 & 62.29 & \textbf{59.13} & {29.68} & \textbf{44.57} & 74.41 & \textbf{54.19} \\

\bottomrule
\end{tabular}
}
% \vspace{-3ex}
\end{table*}

\begin{table*}[t]
\centering
\caption{Ablation study on training data scale.}
% \vspace{-1.5ex}
\label{tab:ab_train_data_size}
\small
\setlength{\tabcolsep}{5pt} % 稍微减小列间距以适应新列
\resizebox{1.0\textwidth}{!}{
\begin{tabular}{ll|cccccc|cccccc|c} % 增加了一个 l 列
\toprule
\multirow{2.5}{*}{\textbf{Metrics}} & \multirow{2.5}{*}{\textbf{Data Scale}} & \multicolumn{6}{c|}{\textbf{Seen}} & \multicolumn{6}{c|}{\textbf{Unseen}} & \multirow{2.5}{*}{\textbf{Avg.}} \\
\cmidrule(lr){3-8} \cmidrule(lr){9-14}
 & & Perfect & Cutout & Dilate & Erode & Merge & Full\_neg & Perfect & Cutout & Dilate & Erode & Merge & Full\_neg & \\
\midrule

% 使用 multirow 合并三行
\multirow{3}{*}{RMSE $\downarrow$} 
 & 25\% & \textbf{0.053} & 0.197 & \textbf{0.247} & \textbf{0.253} & 0.557 & 0.961 & \textbf{0.021} & 0.159 & 0.224 & \textbf{0.222} & 0.489 & 0.953 & 0.361   \\
 & 50\% & 0.160 & 0.184 & 0.270 & 0.277 & 0.383 & 0.756 & 0.122 & 0.152 & 0.236 & 0.228 & 0.383 & 0.760 & 0.326 \\ 
 & \textbf{100\%} & 0.359 & \textbf{0.180} & 0.265 & 0.317 & \textbf{0.334} & \textbf{0.507} & 0.274 & \textbf{0.148} & \textbf{0.179} & 0.263 & \textbf{0.310} & \textbf{0.508} & \textbf{0.303}  \\ \midrule

% 使用 multirow 合并三行
\multirow{3}{*}{$F_2$-M $\uparrow$} 
 & 25\% & \textbf{97.61} & 68.13 & 0.00 & 1.28 & 5.15 & 3.41 & \textbf{98.14} & 84.60 & 0.62 & 0.21 & 4.11 & 4.11 & 30.61 \\
 & 50\% & 93.30 & 75.21 & 11.79 & 5.52 & 60.69 & 27.58 & 95.73 & 89.12 & 11.83 & 7.11 & 41.87 & 37.83 & 46.47 \\ 
 & \textbf{100\%} & 60.58 & \textbf{87.60} & \textbf{69.15} & \textbf{41.62} & \textbf{74.55} & \textbf{69.43} & 64.34 & \textbf{95.18} & \textbf{82.83} & \textbf{34.28} & \textbf{77.73} & \textbf{70.92} & \textbf{69.02} \\ \midrule

% 使用 multirow 合并三行
\multirow{3}{*}{$F_2$-A $\uparrow$} 
 & 25\% & \textbf{97.61} & 38.81 & 1.56 & 3.41 & 4.07 & 6.00 & \textbf{98.14} & 44.64 & 1.44 & 3.06 & 4.91 & 7.06 & 25.89 \\
 & 50\% & 93.30 & 45.52 & 9.68 & 5.62 & 26.64 & 47.84 & 95.73 & 54.38 & 8.81 & 6.68 & 21.92 & 45.85 & 38.50 \\
 & \textbf{100\% }& 60.58 & \textbf{57.57} & \textbf{48.16} & \textbf{32.06} & \textbf{43.00} & \textbf{74.52} & 64.34 & \textbf{62.29} & \textbf{59.13} & \textbf{29.68} & \textbf{44.57} & \textbf{74.41} & \textbf{54.19} \\
\bottomrule
\end{tabular}
}
% \vspace{-2ex}
\end{table*}

\begin{table}[t]
\centering
\vspace{4ex}
\caption{Efficiency comparison between MQ-Auditor and state-of-the-art MLLMs on MQ-RAVSBench. All metrics are reported on a per-mask-sample basis.}

% \vspace{-1.5ex}
\label{tab:efficiency_comparison}
\small
% 定义 X 列为居中对齐，用于数据列
\newcolumntype{Y}{>{\centering\arraybackslash}X}

\begin{tabularx}{\columnwidth}{l YYY}
\toprule
\textbf{Methods} & \textbf{Latency (s) $\downarrow$} & \textbf{Throughput (samples/s) $\uparrow$} & \textbf{Peak Memory (GB)} $\downarrow$ \\
\midrule
Video-LLaMA3-7B~\cite{zhang2025videollama3}     & 7.1   & 0.141 & 31.1 \\
Qwen2.5-Omni-7B~\cite{xu2025qwen2}    & 295.2 & 0.003 & 74.4 \\
Ming-Flash-Omni~\cite{ai2025mingomni}       & 112.8 & 0.009 & 66.0   \\ \midrule
\textbf{MQ-Auditor (ours)}   & \textbf{4.3} & \textbf{0.233} & \textbf{15.1} \\
\bottomrule
\end{tabularx}
% \vspace{-1.5ex}
\end{table}

\section{More Ablation Results}
In this section, we present additional ablation analyses and efficiency comparisons.

\noindent\textbf{Additional Study on Mask Utilization.}
In Sec.~\ref{sec:ablations}, we conduct ablation studies on candidate mask utilization, considering the vanilla mask $\mathcal{M}_t$, the masked frame $\mathcal{V}'_t$, and their combination $[\mathcal{M}_t; \mathcal{V}'_t]$.
Here, we further investigate an alternative variant in which the mask is directly overlaid onto the raw frame using a semi-transparent color (\textit{e.g.}, green), without removing regions outside the mask.
We denote the resulting frame as $\mathcal{\hat{V}}_t$.
The comparison results are reported in Table~\ref{tab:supp_ab_on_mask_utilization}.
Although this overlay-based strategy (`Mask Overlay' in the Table) yields competitive performance for the \textit{Perfect} and \textit{Full\_neg} mask types, it performs noticeably worse on other mask types, including \textit{Merge}.
While $\mathcal{\hat{V}}_t$ highlights the regions selected by the mask, it provides less explicit separation between masked and unmasked regions.
Consequently, we adopt the combination of the binary mask $\mathcal{M}_t$ and the masked frame $\mathcal{V}'_t$ as the default setting, as it offers a better balance between geometric and semantic cues.

% \section{Ablation on the Training Data Scale}
\noindent\textbf{Ablation on the Training Data Scale.}
We explore the impact of training data scale by training our MQ-Auditor with varying sizes of training {video} samples.
As shown in Table~\ref{tab:ab_train_data_size}, using less training data (\textit{e.g.}, 25\%) will increase the proportion of positive perfect samples, improving the performance for \textit{Perfect} mask type.
However, performance for other mask types are significantly destroyed, resulting low $F_2$-M and $F_2$-A scores.
Using the overall performance (`Avg.') as an indicator, the model performance exhibits a clear increase trend using more training data.
This also indicates that it is still non-trivial for developing a model using less training data (\textit{e.g.}, in few-shot or zero-shot settings).

\noindent\textbf{Efficiency Analysis.}
In the main paper, we present extensive experimental results demonstrating the effectiveness and superiority of MQ-Auditor compared with state-of-the-art MLLMs.
Here, we further study its inference efficiency.
Table~\ref{tab:efficiency_comparison} reports per-mask-sample latency, throughput (samples/s), and peak GPU memory usage for MQ-Auditor and three open-source MLLM baselines.
MQ-Auditor achieves the best overall efficiency: it processes a single mask with a latency of 4.3~s, a throughput of 0.233~samples/s, and a peak memory footprint of 15.1~GB.
Compared with the strongest baseline in terms of latency, Video-LLaMA3-7B~\cite{zhang2025videollama3} (7.1~s), MQ-Auditor is approximately $1.6\times$ faster (about 39\% lower latency), delivers around $1.65\times$ higher throughput, and requires about 51\% less peak memory (15.1 vs.\ 31.1~GB).
When compared with larger omni-modal models, the efficiency gains are even more pronounced.
These results further highlight the practical advantages of MQ-Auditor for real-world system deployment.

% These efficiency advantages make MQ-Auditor practical to run as an online audit stage in segmentation pipelines: its low latency and high throughput enable near real-time per-mask assessment at substantially lower hardware cost, and the smaller memory footprint eases multi-task or batch deployments (\textit{e.g.}, auditing many masks concurrently or running on more modest GPUs).

\section{Qualitative Analysis}\label{sec:qualitative_analysis}
\begin{figure}[t]
  \centering
\includegraphics[width=\columnwidth]{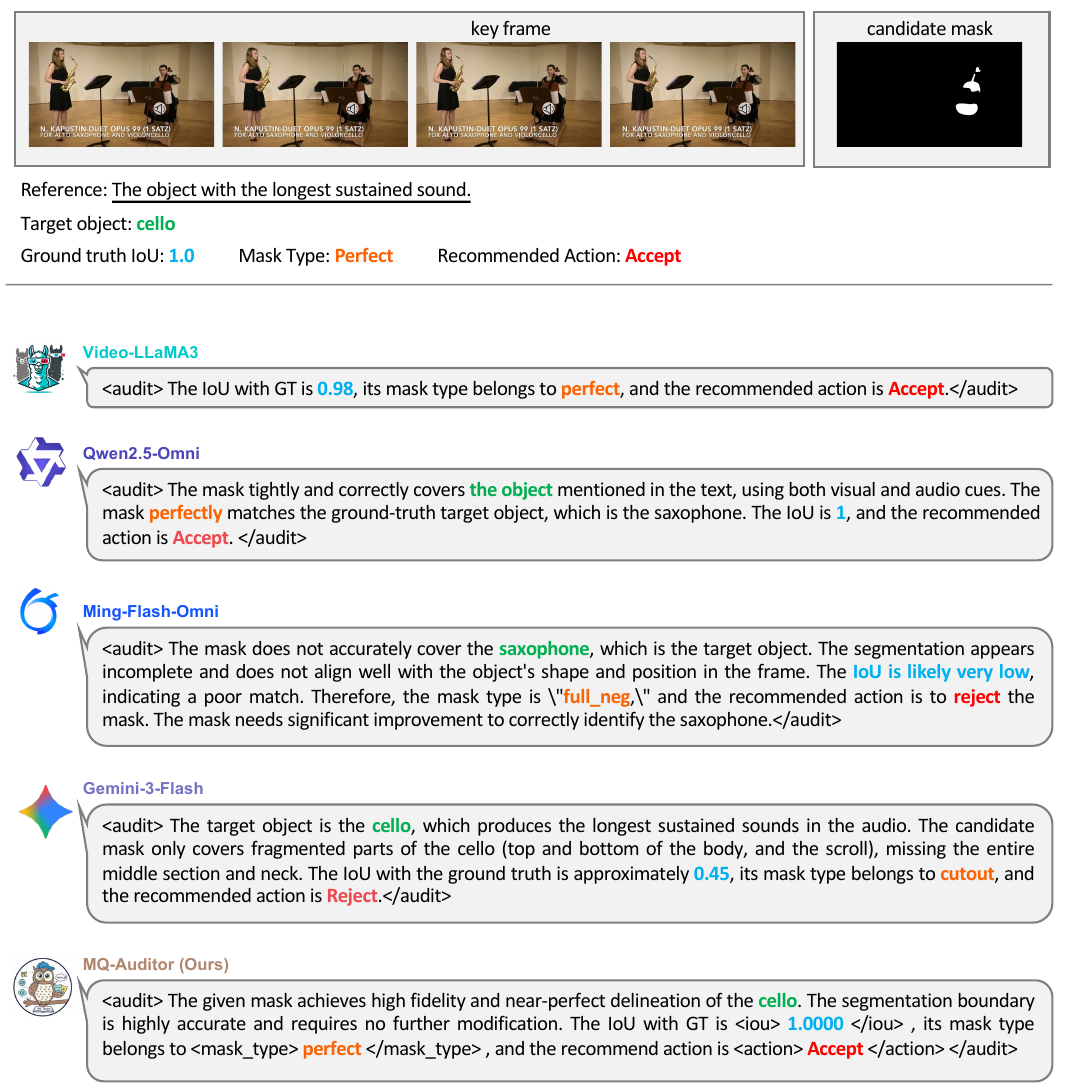}
% \vspace{-3.5ex}
   \caption{Qualitative comparison of different mask quality assessment approaches. Mask type: Perfect.}
   \label{fig:audit_comparison_sample_1}
\end{figure}

\noindent\textbf{Qualitative Comparison.}
Table~\ref{tab:image_based_evaluation_main} in the main paper presents a quantitative comparison between MQ-Auditor and several state-of-the-art MLLMs for mask quality assessment.
To provide more intuitive insights, we further include qualitative comparisons in Figs.~\ref{fig:audit_comparison_sample_1}$\sim$\ref{fig:audit_comparison_sample_6}, where each example corresponds to a specific mask type (\textit{perfect, cutout, dilate, erode, merge,} and \textit{full\_neg}).
The qualitative observations are consistent with the quantitative results.
Video-LLaMA3~\cite{zhang2025videollama3} and Qwen2.5-Omni~\cite{xu2025qwen25omni} tend to accept most candidate masks.
Compared with Video-LLaMA3, Qwen2.5-Omni exhibits better target object recognition, but still fails to accurately estimate IoU or correctly identify mask types.
Ming-Flash-Omni~\cite{ai2025mingomni}, in contrast, shows an overly conservative behavior and tends to reject candidate masks regardless of their actual quality.
Among the compared open-source MLLMs, Gemini-3-Flash~\cite{google2025gemini3} demonstrates stronger overall performance and is able to identify the target object in most cases; however, its IoU estimation is less accurate than that of MQ-Auditor.
These qualitative results further highlight the advantages of MQ-Auditor as a reliable and open-source solution for mask quality assessment.

% \vspace{3ex}

% \textbf{Practical Use of MQ-Auditor.}
\noindent\textbf{Segmentation Improvement via MQ-Auditor.}
Table~\ref{tab:exp_before_and_after_auditor} in the main paper demonstrates that MQ-Auditor can be integrated with prior Ref-AVS models, including EEMC~\cite{wang2024ref} and TGS-Agent~\cite{zhou2025think}, to improve their segmentation performance.
We additionally present qualitative examples to illustrate this practical benefit.
As shown in Fig.~\ref{fig:eemc_before_after_audit_sample_1}(a), the target object is the \textit{flute}, but EEMC incorrectly segments the \textit{piano}.
MQ-Auditor successfully audits the generated mask, identifies the error, and provides the correct target object information.
Based on this audit feedback, segmentation masks are re-generated using Grounded-SAM2~\cite{ren2024grounded}, resulting in accurate segmentation of the target object \textit{flute}.
Similar improvements can be observed in Fig.~\ref{fig:eemc_before_after_audit_sample_2}, where MQ-Auditor again enables effective error diagnosis and mask refinement.
These examples indicate that MQ-Auditor can robustly handle real segmentation outputs produced by existing models and timely identify mask errors for subsequent revision.

% \vspace{3ex}
\noindent\textbf{Failure Cases Analysis.}
Although both quantitative and qualitative results demonstrate the effectiveness of MQ-Auditor, certain failure cases remain.
We discuss two representative examples.
In Fig.~\ref{fig:failure_cases}(a), MQ-Auditor correctly identifies the target object \textit{dog}, but misclassifies an \textit{Erode} mask as \textit{Perfect}, leading to an incorrect recommended action.
Since all pixels in an \textit{Erode} mask belong to the target object, such cases are particularly challenging and can mislead mask quality assessment.
This observation is consistent with our quantitative results in Tables~\ref{tab:image_based_evaluation_main} and~\ref{tab:video_based_evaluation_main}, which show that the \textit{Erode} type is more difficult than other mask types.
Fig.~\ref{fig:failure_cases}(b) presents another failure scenario, where MQ-Auditor produces a reasonable assessment of mask quality but incorrectly understands the semantic identity of the target object.
Such hallucination errors sometimes occur especially under long-context reasoning.
Incorporating reinforcement learning or consistency-based training strategies may help improve alignment between object reasoning and mask quality assessment in future work.

% \vspace{3ex}
\section{Discussion on Limitation}\label{sec:limitation}
In this work, we explore a novel and practical problem of mask quality assessment (MQA).
We demonstrate the benefits of MQA by showing that the proposed model, MQ-Auditor, can serve as an automatic segmentation mask quality rater that estimates IoU without requiring access to ground-truth masks during inference.
Moreover, MQ-Auditor can be integrated with existing Ref-AVS segmentation models to identify potential errors and further improve overall segmentation performance.
Despite these advantages, our work has several limitations.
As discussed in Sec.~\ref{sec:qualitative_analysis}, MQ-Auditor still exhibits failure cases under certain scenarios.
In addition, MQ-RAVSBench is constructed by defining six representative mask types that mimic common error patterns in human annotation and model predictions, covering both geometric and semantic quality issues.
However, segmentation masks produced by human annotators or real-world models can be significantly more complex than these predefined cases, making it difficult to exhaustively enumerate all possible failure modes within a single benchmark.
Nevertheless, MQ-RAVSBench provides a controllable and measurable testbed for initial exploration of mask quality assessment.
We hope this work inspires future research in several directions, including constructing more diverse datasets, developing auditor models with reinforcement learning or improved zero-shot generalization, and extending mask quality assessment to other segmentation settings and tasks.

% \clearpage
\section{Prompts Used in Our Work}\label{sec:prompts}

\subsection{Prompt for \textit{Full\_neg} Mask Construction}\label{sec:prompt_for_full_neg}
As introduced in Sec.~\ref{sec:mask_iou_action}, the construction of \textit{full\_neg} masks is guided by Qwen2.5-VL-72B-Instruct-AWQ~\cite{bai2025qwen25vl}, which is used to generate negative objects that differ from the target referred object.
We provide the detailed prompt below.

\begin{mybox}{}
You are an advanced general visual analysis system capable of precise object detection and reasoning.\\
\vspace{1ex}

\textbf{User Query Target:} \textit{``\{target\_object\}"}
\vspace{2ex}

\textbf{Task:}Analyze the image and detect distinct objects, strictly \textbf{EXCLUDING} the target object mentioned above. Return a JSON list sorted by the specific logic below. \\

\noindent \textbf{Object Granularity (Whole Entities):}
\begin{itemize}[leftmargin=1.5em, nosep]
    \item \textbf{Strictly Whole Objects:} Identify complete entities (\textit{e.g.}, ``person'', ``cat'') rather than parts (\textit{e.g.}, do NOT output ``hand'', ``wheel'').
    \item \textbf{Interactive Context:} If a person or animal is interacting with the target, identify the \textbf{subject} (\textit{e.g.}, ``musician''), not the body part.
\end{itemize}
\vspace{2ex}
\noindent \textbf{Quantity Constraint (Strict):}
\begin{itemize}[leftmargin=1.5em, nosep]
    \item \textbf{Range:} Output between \textbf{2 to 5} objects total.
    \item \textbf{Adaptability:} If the scene is simple, list 2--3 items. If complex, list up to 5.
\end{itemize}
\vspace{2ex}
\noindent \textbf{Sorting Logic (Strict Priority Order):}
\begin{enumerate}[leftmargin=1.5em, nosep]
    \item \textbf{Potential Sound Sources:} Humans, animals, vehicles, electronics, or instruments.
    \item \textbf{Interaction \& Contextual Relevance:} Objects physically/functionally related to the \textit{``\{target\_object\}"} (\textit{e.g.}, leashes, roads).
    \item \textbf{Visual Salience:} Other visually prominent objects.
\end{enumerate}
\vspace{2ex}
\noindent \textbf{Differentiation Requirements:} \\
Use descriptors (color, location) to distinguish similar objects (\textit{e.g.}, ``white van in distance''). \\

\noindent \textbf{Output Format:} \\
Return \textbf{ONLY} a raw JSON list of strings. Do not use Markdown formatting.
\end{mybox}

\subsection{System Prompt for MQ-Auditor}\label{sec:prompt_system_instrution}
In Sec.~\ref{sec:model} -- ``Network'', we introduce how the multimodal inputs are embedded and mention they are sent to a predefined system prompt, which is shown below:

\begin{mybox}{}
video:$<$video\_start$>$$<$video$>$$<$video\_end$>$ \\
audio:$<$audio\_start$>$$<$audio$>$$<$audio\_end$>$ \\
Given the referential expression: \{\textit{reference text}\}, the key frame $<$image\_start$>$$<$image$>$$<$image\_end$>$ and its segmentation mask $<$mask\_start$>$$<$mask$>$$<$mask\_end$>$, please audit the mask quality.
\end{mybox}

Here, the $<$video\_start$>$, $<$video\_end$>$, $<$audio\_start$>$, $<$audio\_end$>$, $<$image\_start$>$, $<$image\_end$>$, $<$mask\_start$>$, $<$mask\_end$>$ are fixed special text tokens.
While the $<$video$>$, $<$audio$>$, $<$image$>$, and $<$mask$>$ will be replaced by real multimodal embeddings of video $\mathcal{V}$, audio $\mathcal{A}$, key frame $\mathcal{V}_t$, and the concatenated embedding of candidate mask $\mathcal{M}_t$ and masked frame $\mathcal{V}'_t$. The \textit{\{reference text\}} will also be replaced by real reference sentence.

% \clearpage
\subsection{Instruction-tuning Prompt for MQ-Auditor}\label{sec:prompt_llm_instruction}
Fig.~\ref{fig:auditor_model} displays an output for the type of \textit{merge} masks that MQ-Auditor is expected to learn.
Here, we provide detailed prompts of each mask type that are used for the instruction tuning of our MQ-Auditor.
Some mask types may have more than one recommended actions and corresponds to different prompts.
The $<$audit$>$, $<$/audit$>$, $<$iou$>$, $<$/iou$>$, $<$mask\_type$>$, $<$/mask\_type$>$, $<$action$>$, and $<$/action$>$ are fixed special text tokens.
The \{iou\_value\} will be replaced by the mask's actual IoU value rounded to four decimal places.
\{Target Object\} will be replaced by the ground truth object category, and \{Negative Object\} used in \textit{Full\_neg} and \textit{Merge} mask types will be replaced by the short descriptive noun phrases of the negative object. 

% \vspace{5ex}

\begin{mybox}{Mask Type: Perfect}
    % \textbf{Ground Truth IoU:} 1 \\
    \textbf{Recommended Action:} Accept \\
    \textbf{Auditing Prompt:} $<$audit$>$ The given mask achieves high fidelity and near-perfect delineation of the \{Target Object\}. The segmentation boundary is highly accurate and requires no further modification. The IoU with GT is $<$iou$>$ \{iou\_value\} $<$/iou$>$, its mask type belongs to $<$mask\_type$>$ perfect $<$/mask\_type$>$, and the recommend action is $<$action$>$ Accept $<$/action$>$ $<$/audit$>$
\end{mybox}

% \vspace{5ex}

\begin{mybox}{Mask Type: Full\_Neg}
    \textbf{Recommended Action:} Reject \\
    \textbf{Auditing Prompt:} $<$audit$>$ The given mask represents a zero overlap error, having erroneously segmented a completely different object, the \{Negative Object\}, instead of the \{Target Object\}. The IoU with GT is $<$iou$>$ \{iou\_value\} $<$/iou$>$, its mask type belongs to $<$mask\_type$>$ full\_neg $<$/mask\_type$>$, and the recommend action is $<$action$>$ Reject $<$/action$>$ $<$/audit$>$
\end{mybox}

% \vspace{ex}

\begin{mybox}{Mask Type: Cutout}
    \textbf{Recommended Action:} Minor Revision \\
    \textbf{Auditing Prompt:} $<$audit$>$ The given mask correctly delineates the \{Target Object\} but suffers from a false negative error with minor internal defects, resulting in partial loss of foreground connectivity. The IoU with GT is $<$iou$>$ \{iou\_value\} $<$/iou$>$, its mask type belongs to $<$mask\_type$>$ cutout $<$/mask\_type$>$, and the recommend action is $<$action$>$ Minor Revision $<$/action$>$ $<$/audit$>$ \\
    \rule{\linewidth}{0.4pt} \\
    \textbf{Recommended Action:} Major Revision \\
    \textbf{Auditing Prompt:} $<$audit$>$ The given mask delineates the \{Target Object\} but suffers from severe internal false negative errors, resulting in a significant loss of foreground pixels and a low coverage area. The IoU with GT is $<$iou$>$ \{iou\_value\} $<$/iou$>$, its mask type belongs to $<$mask\_type$>$ cutout $<$/mask\_type$>$, and the recommend action is $<$action$>$ Major Revision $<$/action$>$ $<$/audit$>$
\end{mybox}

% \vspace{5ex}

\begin{mybox}{Mask Type: Dilate}
    \textbf{Recommended Action:} Minor Revision \\
    \textbf{Auditing Prompt:} $<$audit$>$ The given mask correctly delineates the \{Target Object\} but suffers from a false positive error due to slight boundary expansion, incorrectly including minor background pixels. The IoU with GT is $<$iou$>$ \{iou\_value\} $<$/iou$>$, its mask type belongs to $<$mask\_type$>$ dilate $<$/mask\_type$>$, and the recommend action is $<$action$>$ Minor Revision $<$/action$>$ $<$/audit$>$ \\
    \rule{\linewidth}{0.4pt} \\
    \textbf{Recommended Action:} Major Revision \\
    \textbf{Auditing Prompt:} $<$audit$>$ The given mask delineates the \{Target Object\} but suffers from severe boundary expansion, resulting in a substantial false positive error by covering a large area of the background. The IoU with GT is $<$iou$>$ \{iou\_value\} $<$/iou$>$, its mask type belongs to $<$mask\_type$>$ dilate $<$/mask\_type$>$, and the recommend action is $<$action$>$ Major Revision $<$/action$>$ $<$/audit$>$
\end{mybox}

\vspace{2ex}

\begin{mybox}{Mask Type: Erode}
    \textbf{Recommended Action:} Minor Revision \\
    \textbf{Auditing Prompt:} $<$audit$>$ The given mask correctly delineates the \{Target Object\} but suffers from a false negative error due to slight boundary shrinkage, causing minor loss of edge detail. The IoU with GT is $<$iou$>$ \{iou\_value\} $<$/iou$>$, its mask type belongs to $<$mask\_type$>$ erode $<$/mask\_type$>$, and the recommend action is $<$action$>$ Minor Revision $<$/action$>$ $<$/audit$>$ \\
    \rule{\linewidth}{0.4pt} \\
    \textbf{Recommended Action:} Major Revision \\
    \textbf{Auditing Prompt:} $<$audit$>$ The given mask delineates the \{Target Object\} but suffers from severe boundary shrinkage, leading to a significant false negative error and a highly inaccurate final shape. The IoU with GT is $<$iou$>$ \{iou\_value\} $<$/iou$>$, its mask type belongs to $<$mask\_type$>$ erode $<$/mask\_type$>$, and the recommend action is $<$action$>$ Major Revision $<$/action$>$ $<$/audit$>$
\end{mybox}

\vspace{2ex}

\begin{mybox}{Mask Type: Merge}
    \textbf{Recommended Action:} Minor Revision \\
    \textbf{Auditing Prompt:} $<$audit$>$ The given mask correctly delineates the \{Target Object\} but contains partial errors by including a portion of the \{Negative Object\}. The IoU with GT is $<$iou$>$ \{iou\_value\} $<$/iou$>$, its mask type belongs to $<$mask\_type$>$ merge $<$/mask\_type$>$, and the recommend action is $<$action$>$ Minor Revision $<$/action$>$ $<$/audit$>$ \\
    \rule{\linewidth}{0.4pt} \\
    \textbf{Recommended Action:} Major Revision \\
    \textbf{Auditing Prompt:} $<$audit$>$ The given mask correctly captures the \{Target Object\} but also erroneously merges a significant portion of the \{Negative Object\}. The IoU with GT is $<$iou$>$ \{iou\_value\} $<$/iou$>$, its mask type belongs to $<$mask\_type$>$ merge $<$/mask\_type$>$, and the recommend action is $<$action$>$ Major Revision $<$/action$>$ $<$/audit$>$ \\
    \rule{\linewidth}{0.4pt} \\
    \textbf{Recommended Action:} Reject \\
    \textbf{Auditing Prompt:} $<$audit$>$ The given mask is highly erroneous, segmenting the \{Target Object\} but merging an excessively large, non-target \{Negative Object\}, resulting in a quite low IoU. The IoU with GT is $<$iou$>$ \{iou\_value\} $<$/iou$>$, its mask type belongs to $<$mask\_type$>$ merge $<$/mask\_type$>$, and the recommend action is $<$action$>$ Reject $<$/action$>$ $<$/audit$>$
\end{mybox}

\subsection{Prompt for Evaluating Other MLLMs}\label{sec:prompt_other_mllm_eval}
Table~\ref{tab:image_based_evaluation_main} provides a comparison between our MQ-Auditor with several powerful open-source and closed-source MLLMs. We list the detailed prompts used for evaluating these MLLMs.

\begin{mybox}{}
\textbf{Text Query (target object to segment):} \\
\{reference text\} \\
Given video, audio, a reference video frame where the target object appears, a candidate binary segmentation mask for that reference frame, your job is to evaluate the candidate mask quality for the specified target object, 
and output exactly one $<$audit$>$ ... $<$/audit$>$ block.

\vspace{1em}
\textbf{Audio:} \{real audio data\} \\
\textbf{Video:} \{real video data\} \\
\textbf{Key video frame:} \{real key frame\} \\
\textbf{Binary mask:} \{real binary mask\}

\vspace{1em}
You must reason about whether the mask tightly and correctly covers the 
object mentioned in the text, using both visual and audio cues if helpful.

There are 6 possible mask types (must use exactly one of these keywords): \\
\textbf{Perfect:} The mask perfectly matches the ground-truth target object. IoU=1, action: Accept \\
\textbf{Cutout/Dilate/Erode:} The mask covers the correct target object but has missing/outer/shrunk regions. 
IoU in [0.85, 0.90), action: Minor Revision. IoU in [0.75, 0.80), action: Major Revision. \\
\textbf{Merge:} The mask includes the correct object but also merges in extra regions from unrelated objects.\\
\hspace*{1.5em}- Small merge: IoU in [0.90, 1.0), action: Minor Revision. \\
\hspace*{1.5em}- Medium merge: IoU in [0.75, 0.90), action: Major Revision. \\
\hspace*{1.5em}- Large merge: IoU in [0.0, 0.75), action: Reject. \\
\textbf{Full\_neg:} The mask does not include any pixels of the target object at all. IoU = 0.0. action: Reject.

\vspace{1em}
You must output a single $<$audit$>$ block with the following structure: \\
$<$audit$>$[reasoning on mask quality] The IoU with GT is $<$iou$>$ IOU $<$/iou$>$, its mask type belongs to $<$mask$>$ MASKTYPE $<$/mask$>$, and the recommended action is $<$action$>$ ACTION $<$/action$>$.$<$/audit$>$
\end{mybox}

%%%%%%%%%%%%%%%%%%%%%%%%%%%%%%%%%%%%%%%%%%%%%%%%%%%%%%%%%%%%%%%%%%%%%%%%%%%%%%%%%%%%%%%
% 6 个不同 mask type，各个 MLLMs 的 audit 定性比较
%%%%%%%%%%%%%%%%%%%%%%%%%%%%%%%%%%%%%%%%%%%%%%%%%%%%%%%%%%%%%%%%%%%%%%%%%%%%%%%%%%%%%%%

\begin{figure}[t]
  \centering
\includegraphics[width=\columnwidth]{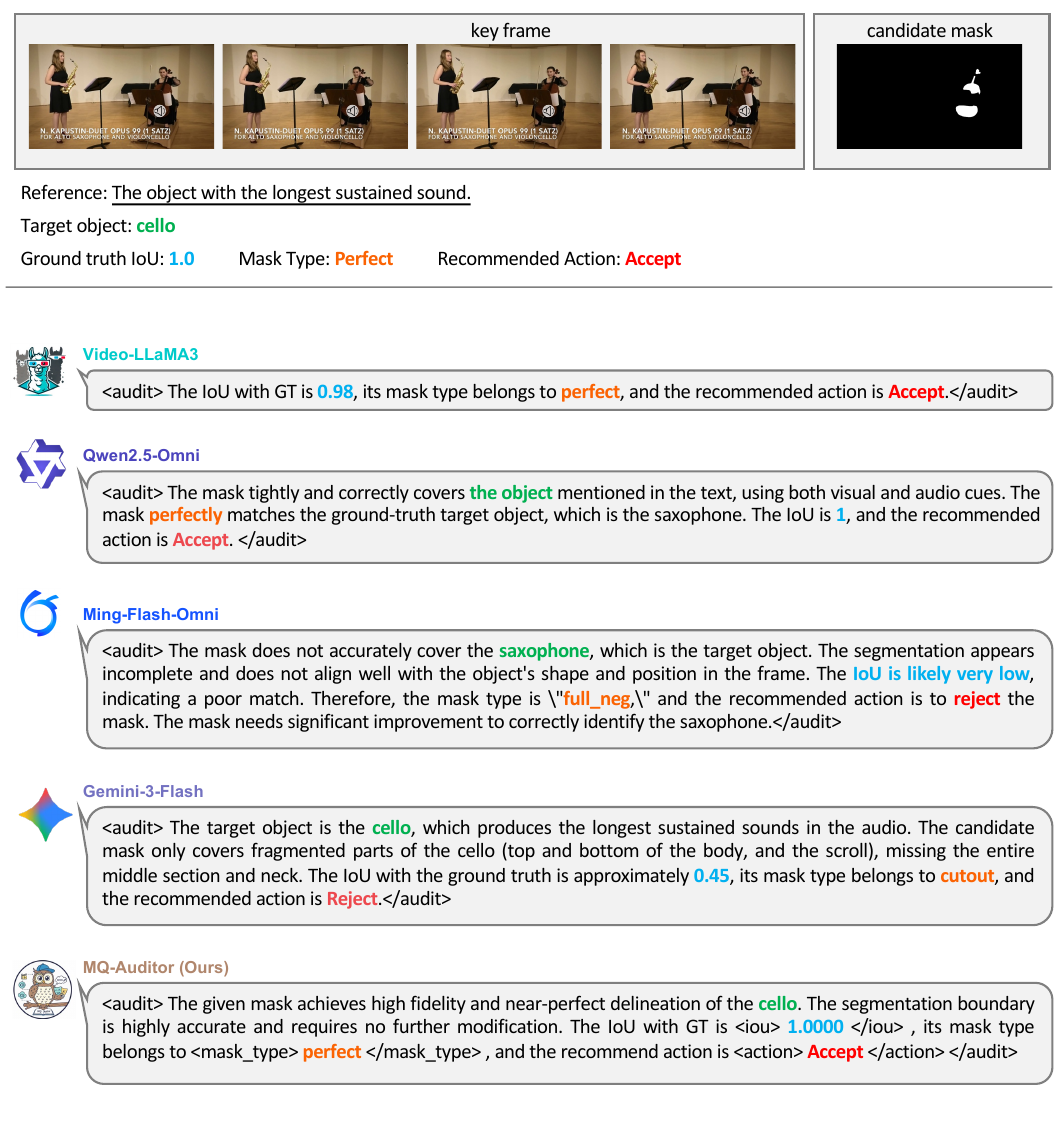}
% \vspace{3.5ex}
   \caption{Qualitative comparison of different mask quality assessment approaches. Mask type: Cutout.}
   \label{fig:audit_comparison_sample_2}
\end{figure}

\begin{figure}[t]
  \centering
\includegraphics[width=\columnwidth]{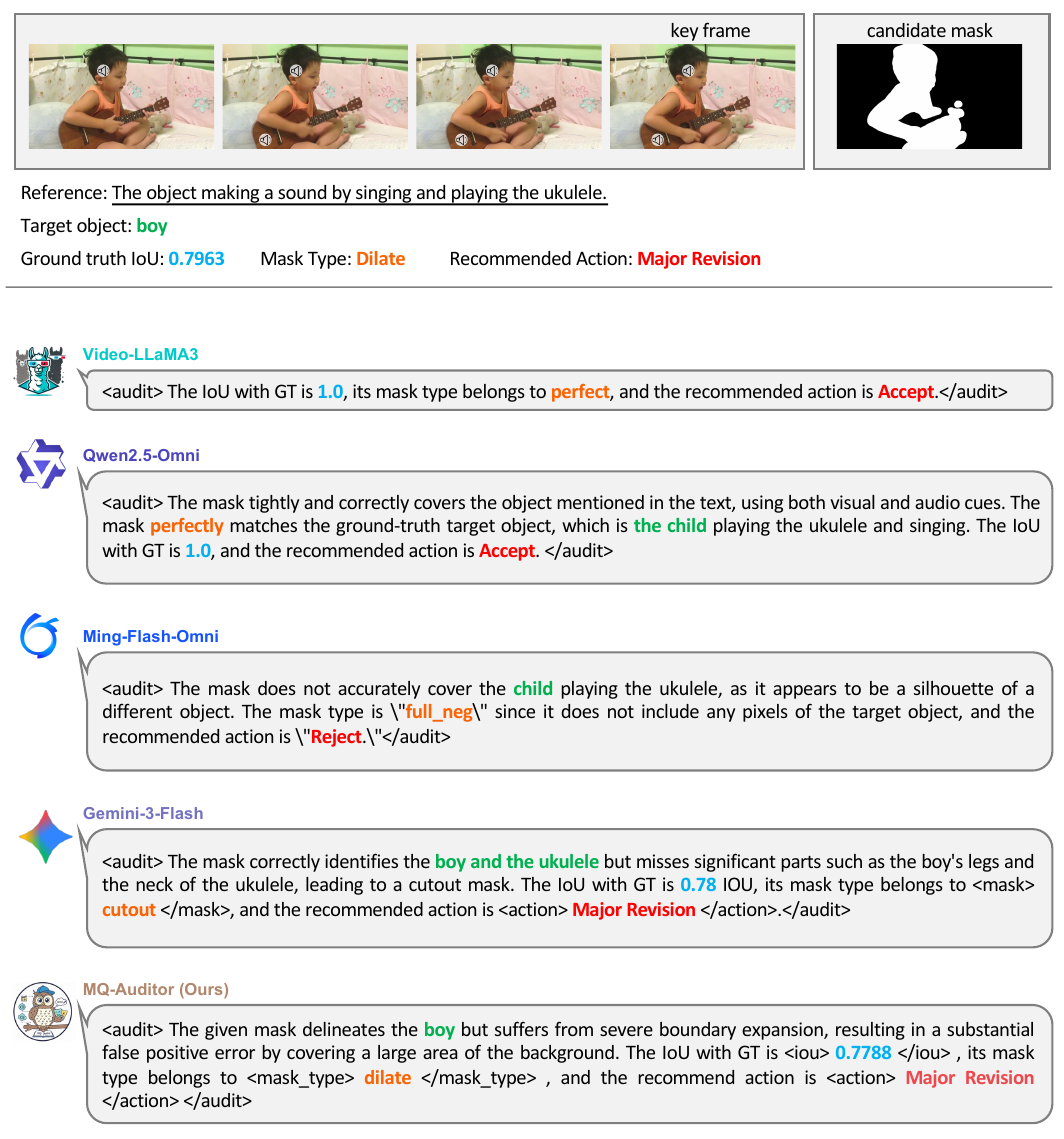}
% \vspace{-3.5ex}
   \caption{Qualitative comparison of different mask quality assessment approaches. Mask type: Dilate.}
   \label{fig:audit_comparison_sample_3}
\end{figure}

\begin{figure}[t]
  \centering
\includegraphics[width=\columnwidth]{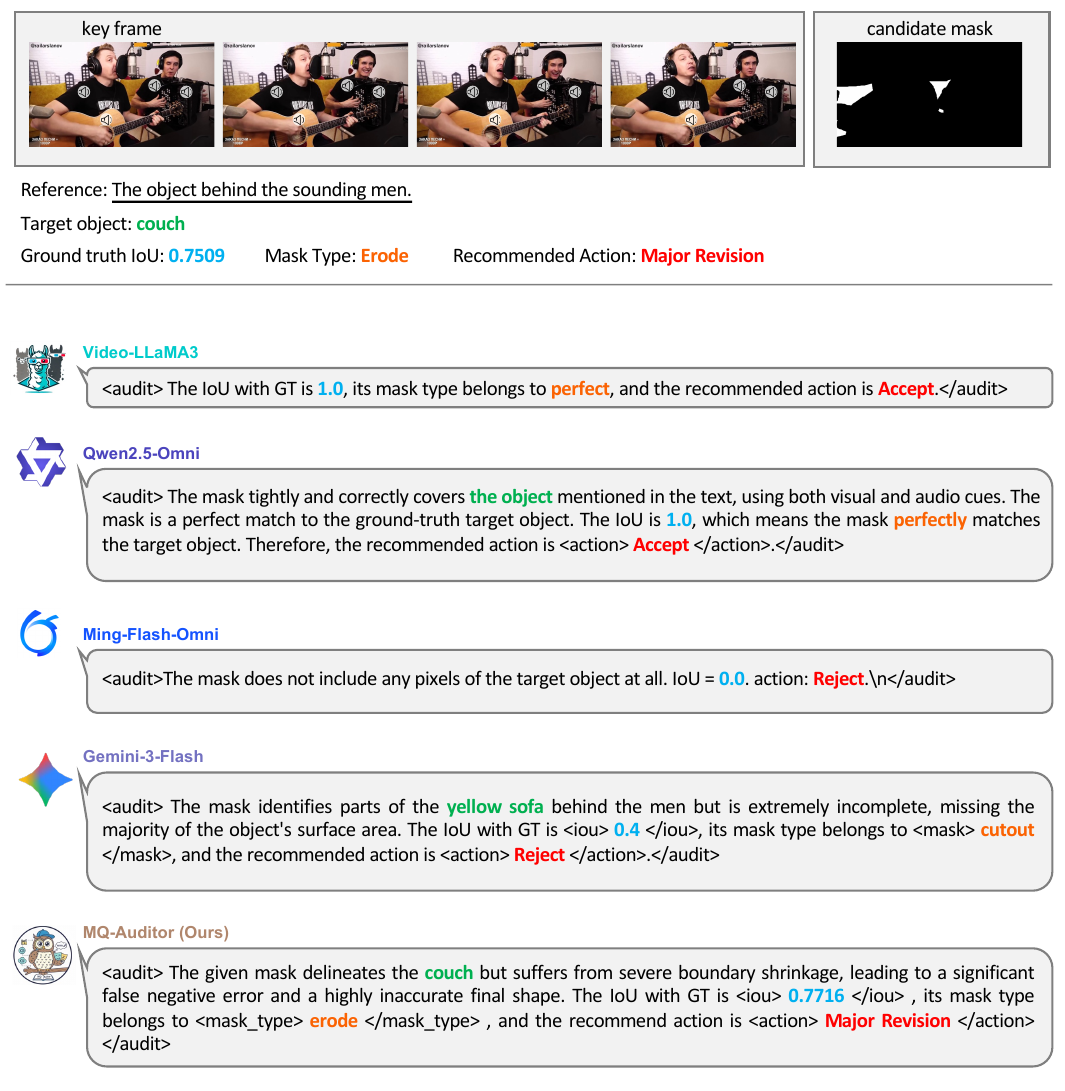}
% \vspace{-3.5ex}
   \caption{Qualitative comparison of different mask quality assessment approaches. Mask type: Erode.}
   \label{fig:audit_comparison_sample_4}
\end{figure}

\begin{figure}[t]
  \centering
\includegraphics[width=\columnwidth]{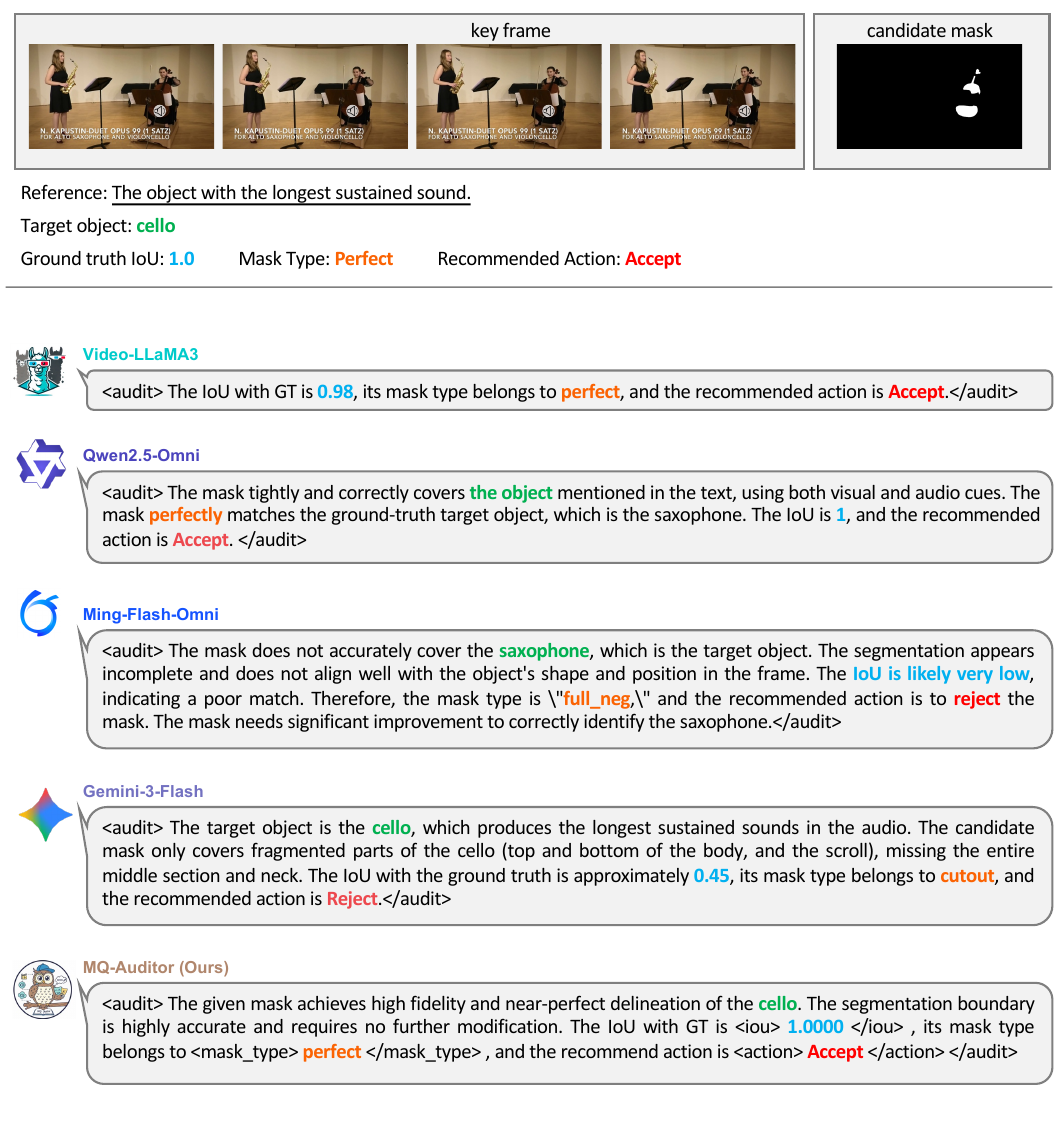}
% \vspace{-3.5ex}
   \caption{Qualitative comparison of different mask quality assessment approaches. Mask type: Merge.}
   \label{fig:audit_comparison_sample_5}
\end{figure}

\begin{figure}[t]
  \centering
\includegraphics[width=\columnwidth]{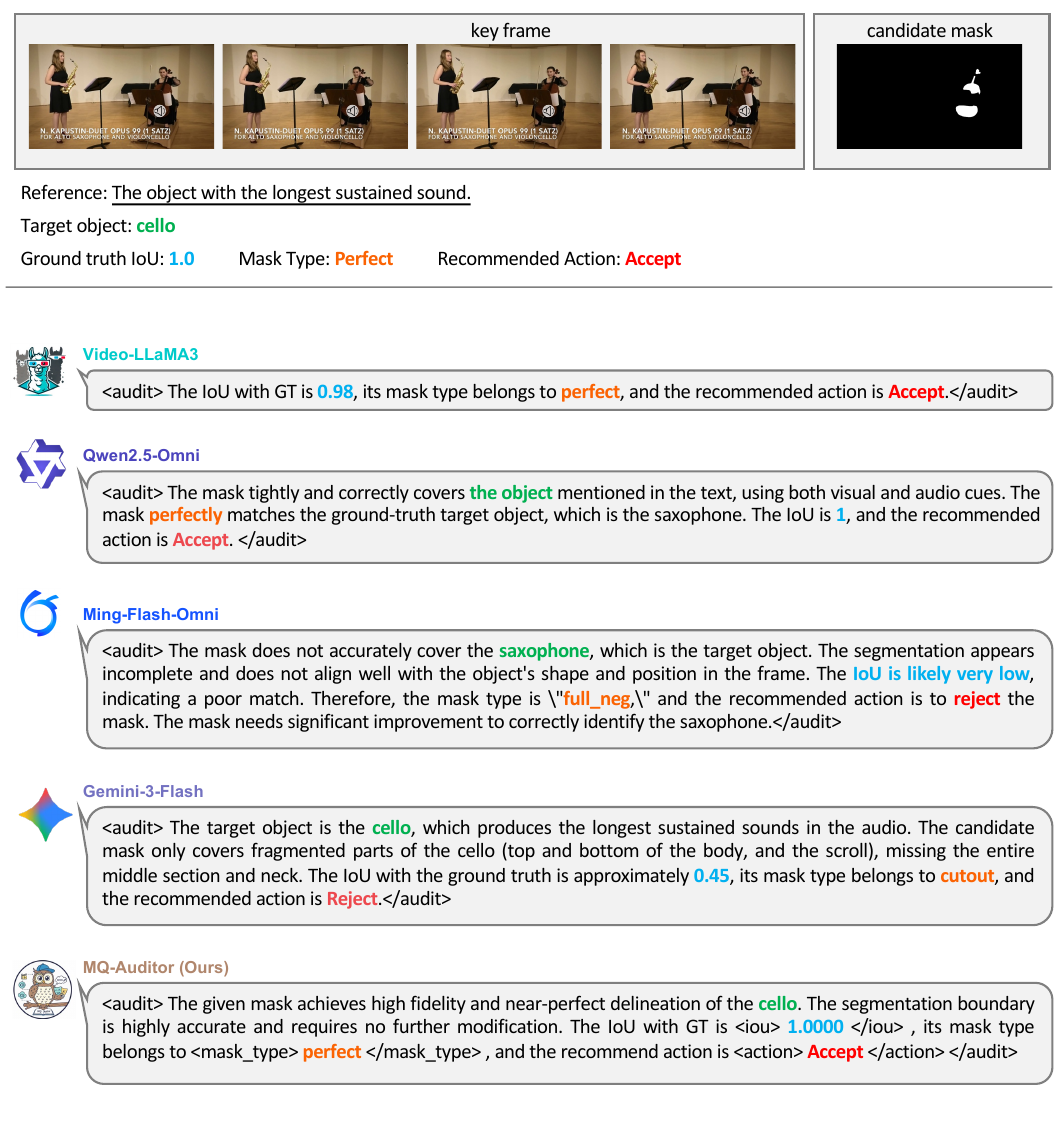}
% \vspace{-3.5ex}
   \caption{Qualitative comparison of different mask quality assessment approaches. Mask type: Full\_neg.}
   \label{fig:audit_comparison_sample_6}
\end{figure}

%%%%%%%%%%%%%%%%%%%%%%%%%%%%%%%%%%%%%%%%%%%%%%%%%%%%%%%%%%%%%%%%%%%%%%%%%%%%%%%%%%%%%%%
% EEMC/TGS-Agent before and after MQ-Auditor
%%%%%%%%%%%%%%%%%%%%%%%%%%%%%%%%%%%%%%%%%%%%%%%%%%%%%%%%%%%%%%%%%%%%%%%%%%%%%%%%%%%%%%%

\begin{figure}[t]
  \centering
\includegraphics[width=\columnwidth]{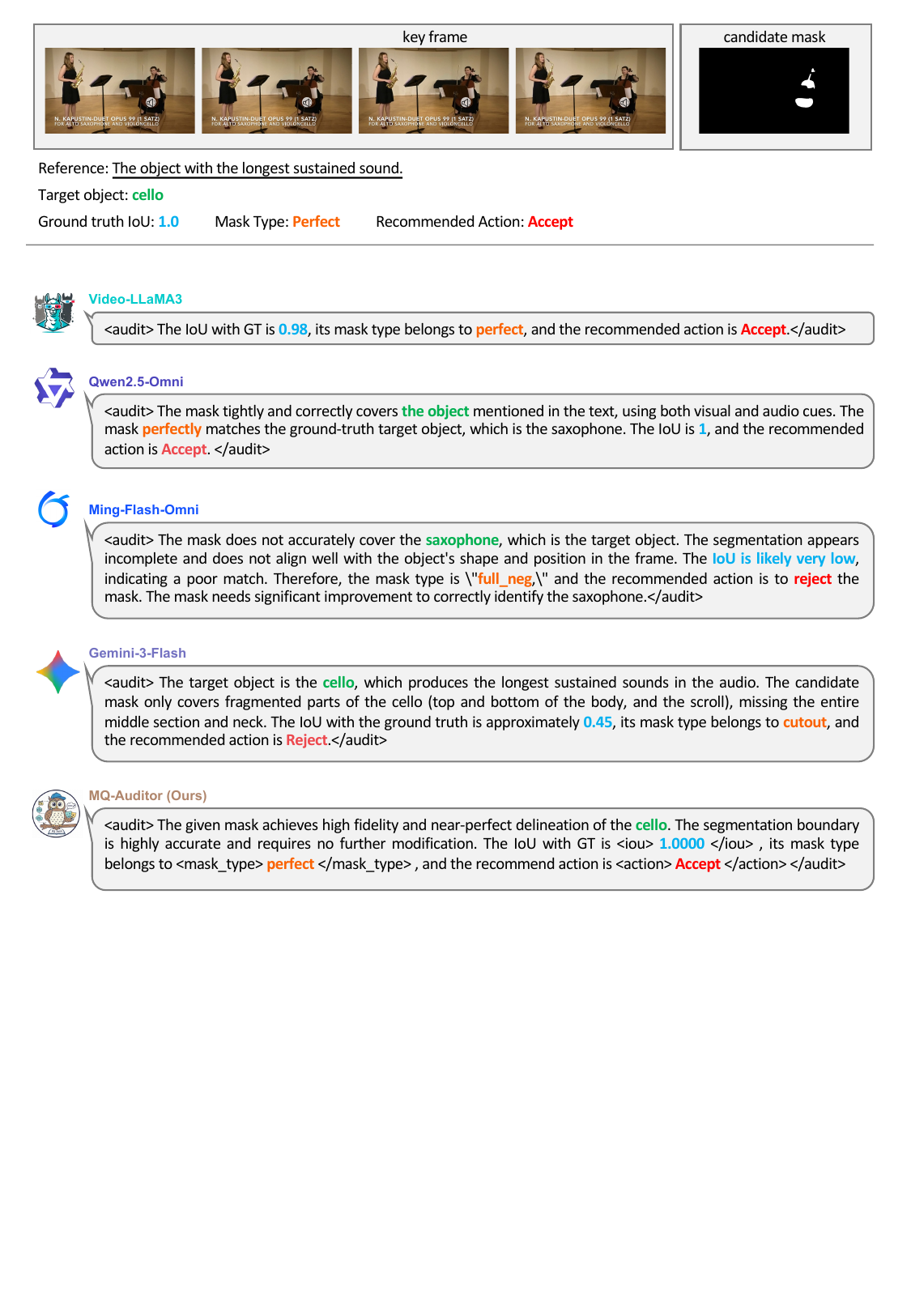}
% \vspace{-3.5ex}
   \caption{Segmentation performance comparison before and after using our MQ-Auditor. The prior state-of-the-art Ref-AVS method, EEMC~\cite{wang2024ref}, is used in these examples. Our MQ-Auditor can assess the quality of masks generated by real Ref-AVS models and provides correct target object information to guide mask refinement.}
   \label{fig:eemc_before_after_audit_sample_1}
\end{figure}

\begin{figure}[t]
  \centering
\includegraphics[width=\columnwidth]{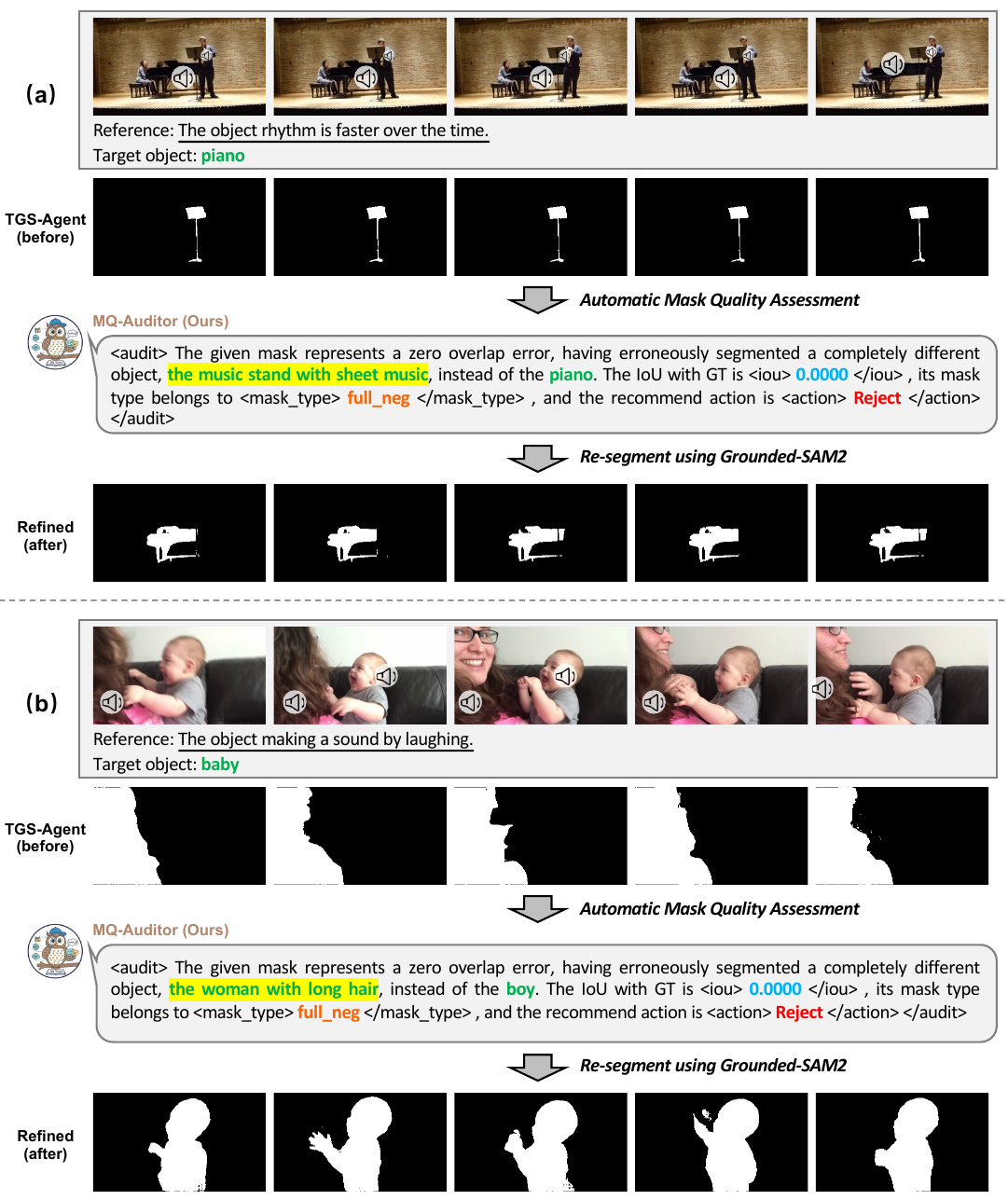}
% \vspace{-3.5ex}
\caption{Segmentation performance comparison before and after applying MQ-Auditor. The prior state-of-the-art Ref-AVS method TGS-Agent~\cite{zhou2025think} is used in these examples. Our MQ-Auditor can assess the quality of masks generated by real Ref-AVS models and provides correct target object information to guide mask refinement.}

   \label{fig:eemc_before_after_audit_sample_2}
\end{figure}

%%%%%%%%%%%%%%%%%%%%%%%%%%%%%%%%%%%%%%%%%%%%%%%%%%%%%%%%%%%%%%%%%%%%%%%%%%%%%%%%%%%%%%%
% failures cases
%%%%%%%%%%%%%%%%%%%%%%%%%%%%%%%%%%%%%%%%%%%%%%%%%%%%%%%%%%%%%%%%%%%%%%%%%%%%%%%%%%%%%%%
\begin{figure}[t]
  \centering
\includegraphics[width=\columnwidth]{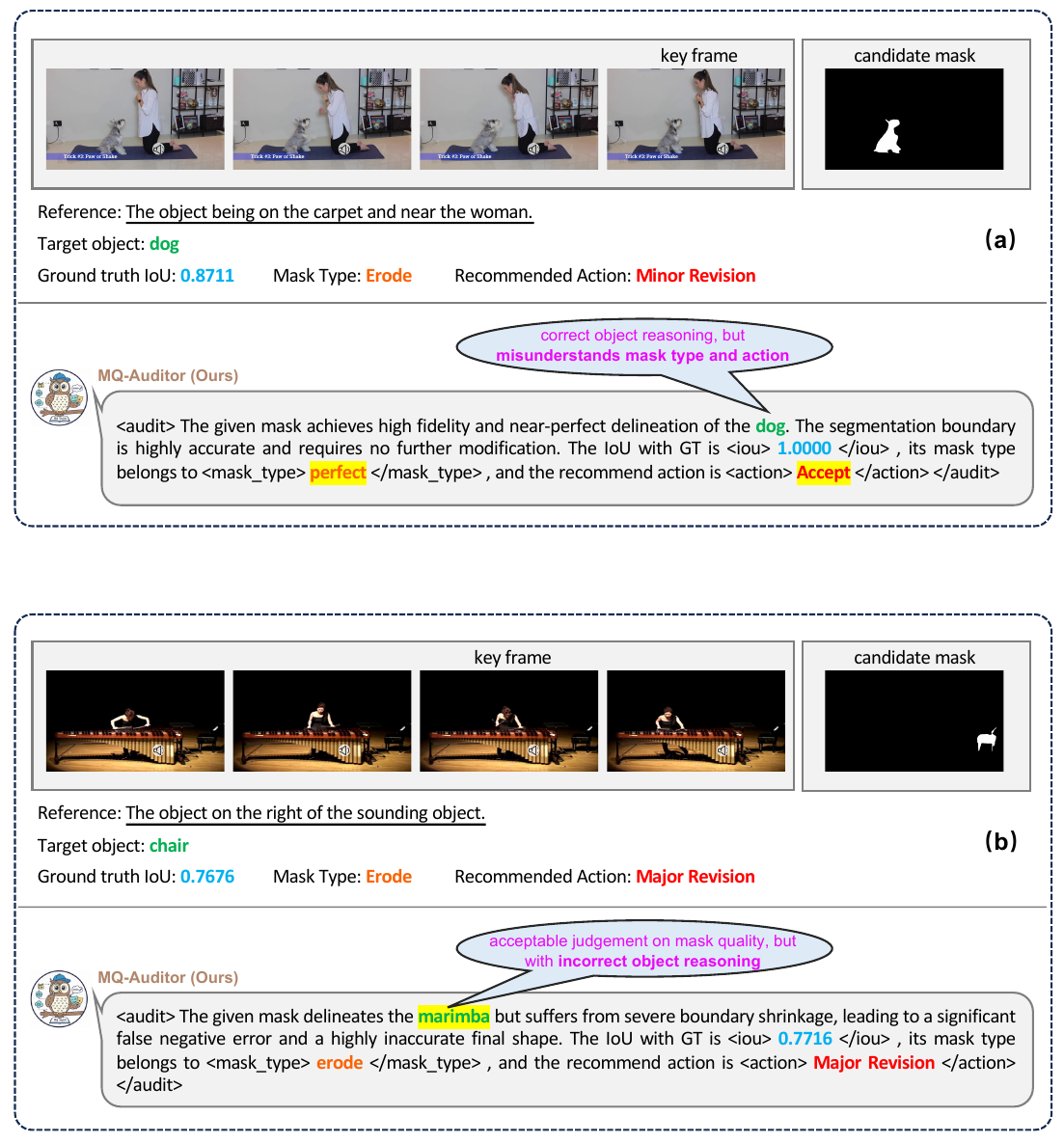}
% \vspace{-3.5ex}
   \caption{Visualization of two representative failure cases. MQ-Auditor may suffer from mask type misclassification or object recognition hallucinations.}
   \label{fig:failure_cases}
\end{figure}

\end{document}